\documentclass[journal]{IEEEtran}
\usepackage{amsmath,amsopn,amssymb}
\usepackage{graphicx,xspace,color,soul}
\usepackage{epsfig,subfigure}
\usepackage{longtable,multirow}
\usepackage{tabularx}
\usepackage{array,float}
\usepackage{cite}
\usepackage{algorithmic}
\usepackage[linesnumbered, ruled]{algorithm2e}
\usepackage{bm,epstopdf}
\usepackage{caption2}

\renewcommand{\vec}[1]{\mathbf{#1}}

\newcolumntype{Y}{>{\centering\arraybackslash}X}

\def\ie{{\textit{i.e.}}}

\begin{document}


\title{3D Dynamic Point Cloud Denoising via Spatial-Temporal Graph Learning}

\author{
    Wei~Hu,~\IEEEmembership{Member,~IEEE,}
	Qianjiang~Hu,~\IEEEmembership{Student Member,~IEEE,}
	Zehua~Wang,~\IEEEmembership{Student Member,~IEEE,}
	and~Xiang~Gao,~\IEEEmembership{Student Member,~IEEE}
	
}


\maketitle

\begin{abstract}

The prevalence of accessible depth sensing and 3D laser scanning techniques has enabled the convenient acquisition of 3D dynamic point clouds, which provide efficient representation of arbitrarily-shaped objects in motion. 
Nevertheless, dynamic point clouds are often perturbed by noise due to hardware, software or other causes. 
While a plethora of methods have been proposed for static point cloud denoising, few efforts are made for the denoising of dynamic point clouds with varying number of irregularly-sampled points in each frame. 
In this paper, we represent dynamic point clouds naturally on graphs and address the denoising problem by inferring the underlying graph via spatio-temporal graph learning, exploiting both the intra-frame similarity and inter-frame consistency. 
Firstly, assuming the availability of a relevant feature vector per node, we pose spatial-temporal graph learning as optimizing a Mahalanobis distance metric $\mathbf{M}$, which is formulated as the minimization of graph Laplacian regularizer. 
Secondly, to ease the optimization of the symmetric and positive definite metric matrix $\mathbf{M}$, we decompose it into $\mathbf{M}=\mathbf{R}^{\top}\mathbf{R}$ and solve $\mathbf{R}$ instead via proximal gradient.
Finally, based on the spatial-temporal graph learning, we formulate dynamic point cloud denoising as the joint optimization of the desired point cloud and underlying spatio-temporal graph, which leverages both intra-frame affinities and inter-frame consistency and is solved via alternating minimization. 
Experimental results show that the proposed method significantly outperforms independent denoising of each frame from state-of-the-art static point cloud denoising approaches.         

\end{abstract}


\vspace{-0.05in}
\section{Introduction}
\label{sec:intro}
\begin{figure}[htbp]
     \centering
     \includegraphics[width=\linewidth]{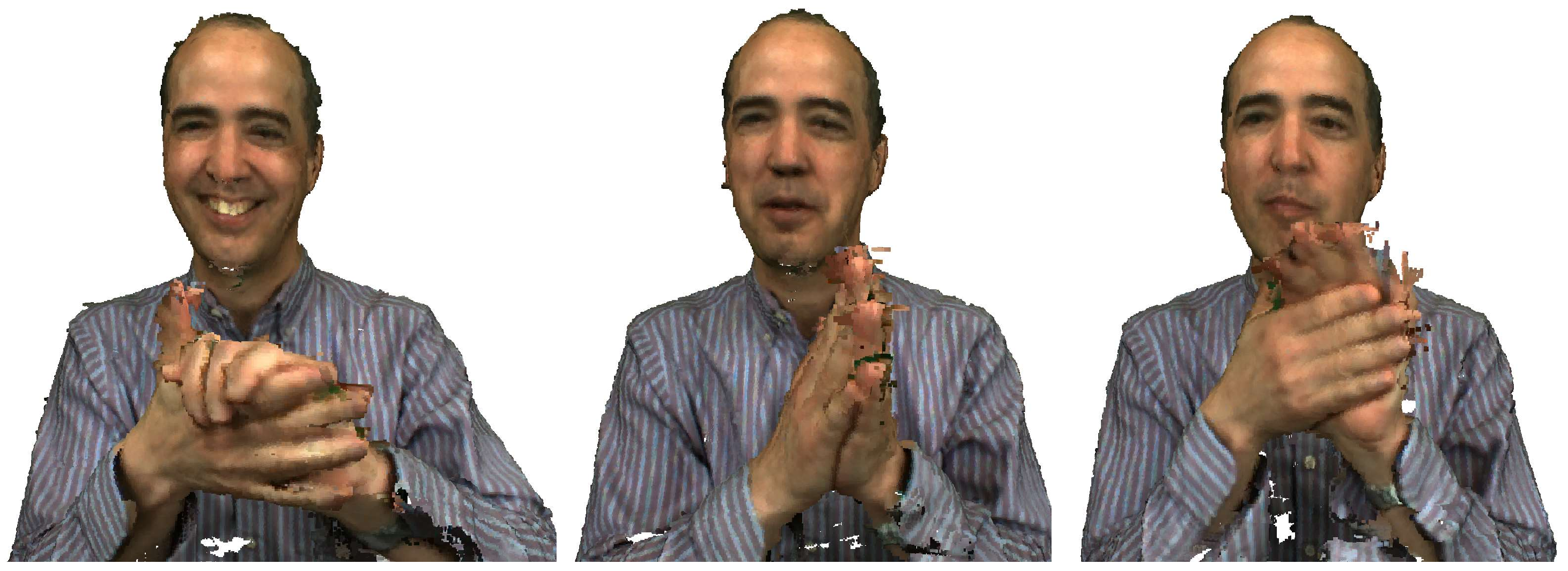}
     \caption{Three frames ($0000$, $0010$, $0020$) in the dynamic point cloud sequence \textit{Phil} \cite{loop2016microsoft}, which is captured by four frontal RGBD cameras. The hands are noisy in particular. }
     \label{fig:noisy}
 \end{figure}

The maturity of depth sensing and 3D laser scanning techniques has enabled convenient acquisition of 3D dynamic point clouds, a natural representation for arbitrarily-shaped objects varying over time \cite{Rusu20113DIH}. A dynamic point cloud consists of a sequence of static point clouds, each of which is composed of a set of points defined on irregular grids, as shown in Fig.~\ref{fig:noisy}. Each point has geometry information (\ie, 3D coordinates) and possibly attribute information such as color. Because of the efficient representation, dynamic point clouds have been widely applied in various fields, such as 3D immersive tele-presence, navigation for autonomous vehicles, gaming and animation \cite{tulvan16}. 

Point clouds are often perturbed by noise, which comes from hardware, software or other causes. Hardware wise, noise occurs due to the inherent limitations of the acquisition equipment, as demonstrated in Fig.~\ref{fig:noisy}. Software wise, in the case of generating point clouds with existing algorithms, points may locate somewhere completely wrong due to imprecise triangulation ({\it e.g.}, a false epipolar matching). 
Due to the irregular sampling and varying number of points in each frame, dynamic point cloud denoising is quite challenging to address. 

However, few efforts are made for the denoising of \textit{dynamic} point clouds in the literature, while many approaches have been proposed for static point cloud denoising. 
Existing denoising methods for static point clouds mainly include moving least squares (MLS)-based methods \cite{Alexa2003Computing,Guennebaud2007Algebraic,A2009Feature}, locally optimal projection (LOP)-based methods \cite{Huang2013Edge,Hui2009Consolidation,Lipman2007Parameterization}, sparsity-based methods \cite{avron,Mattei2017Point}, non-local similarity-based methods \cite{Dinesh2018Fast,zeng20183d}, and graph-based methods \cite{dinesh18arxiv,zeng18arxiv,duan2018weighted,Hu19tsp}. 
Whereas it is possible to apply existing static point cloud denoising methods to each frame of a dynamic point cloud sequence independently, the \textit{inter-frame correlation} would be neglected, which may lead to inconsistent denoising results in the temporal domain. 

To this end, we propose to exploit inter-frame correlation for dynamic point cloud denoising, which \textit{not only enforces the temporal consistency but also provides additional temporal information for denoising}. 
Since point clouds are irregular, we represent dynamic point clouds naturally on {\it spatio-temporal graphs}, where each node represents a point and each edge captures the similarity between a pair of spatially or temporally adjacent points. 
In this paper, we focus on the geometry of point clouds, and thus the corresponding graph signal refers to the 3D Cartesian coordinates of points. 
Further, assuming a feature vector per node is available, we propose to infer the underlying graph representation via \textit{spatial-temporal graph learning}, which is posed as the optimization of a Mahalanobis distance metric \cite{mahalanobis1936}. 
We then reconstruct dynamic point clouds by regularization from the learned graph representation, exploring both the intra-frame affinities and inter-frame correlation.

Specifically, we first propose spatial-temporal graph learning assuming the availability of relevant features $\mathbf{f}_i \in \mathbb{R}^K$ per node $i$.
Since an edge weight characterizes pairwise similarities between nodes, it is common to define an edge weight as a function of the feature distance. 
Given two connected nodes $i$ and $j$ and their feature difference $\mathbf{f}_i - \mathbf{f}_j$, we assume the edge weight $w_{i,j}=\exp \{ - (\mathbf{f}_i - \mathbf{f}_j)^{\top} \mathbf{M} (\mathbf{f}_i - \mathbf{f}_j) \}$, where $\mathbf{M} \in \mathbb{R}^{K \times K}$ is the \textit{Mahalanobis distance} metric matrix \cite{mahalanobis1936}.
Considering a target point cloud signal with $N$ points $\mathbf{z} \in \mathbb{R}^N$, we seek to optimize $\mathbf{M}$ by minimizing the \textit{Graph Laplacian Regularizer} (GLR) \cite{pang2017graph} $\mathbf{z}^{\top} \mathbf{L}(\mathbf{M}) \mathbf{z}$, which measures the smoothness of the signal $\mathbf{z}$ with respect to the graph Laplacian matrix $\mathbf{L}$ \cite{chung1997spectral}.
Further, since $\mathbf{M}$ is symmetric and positive definite, we decompose it as $\mathbf{M}=\mathbf{R}^{\top}\mathbf{R}$, $\mathbf{R} \in \mathbb{R}^{K \times K}$, so as to remove the constraint of positive definiteness for computation efficiency. 
Hence, given each target frame and its previous reconstructed frame in the input dynamic point cloud, we cast spatial-temporal graph learning as the estimation of the distance metrics $\{\mathbf{R}_s,\mathbf{R}_t\}$ from spatially and temporally neighboring points respectively, with $\mathbf{R}_s \in \mathbb{R}^{K \times K}$ as the spatial distance metric and $\mathbf{R}_t \in \mathbb{R}^{K \times K}$ as the temporal distance metric. 
We then propose an algorithm to optimize the distance metrics via proximal gradient descent \cite{Parikh31}. 

Secondly, based on the spatial-temporal graph learning, we formulate dynamic point cloud denoising as the joint optimization of the desired point cloud and underlying spatial-temporal graph representation. 
To exploit the intra-frame correlation, we first divide each frame into overlapping patches. 
Each irregular patch is defined as a local point set consisting of a centering point and its $k$-nearest neighbors. 
For each target patch in the current noisy frame, we search for its most similar patches to exploit the intra-frame correlation, and search for its corresponding patch in the previously reconstructed frame to explore the inter-frame correlation. 
Given the searched patches, we construct spatio-temporal edge connectivities based on patch similarity. 
Then we regularize the formulation by smoothness of similar patches in the current frame via GLR as well as temporal consistency over corresponding patches with respect to the underlying spatio-temporal graph.   

Finally, we propose an efficient algorithm to address the above problem formulation.
We design an alternating minimization algorithm to optimize the underlying frame and spatial-temporal graph alternately. 
When the underlying frame is initialized or fixed, the graph is optimized from the proposed spatio-temporal graph learning. 
When the graph is fixed, we update the underlying frame via a closed-form solution.  
Then we update patch construction, similar patch search and edge connectivities from each update of the underlying frame. 
The process is iterated until convergence. 
Experimental results show that the proposed method outperforms independent denoising of each frame from state-of-the-art static point cloud denoising approaches on nine widely employed dynamic point cloud sequences.

In summary, the main contributions of our work include:
\begin{itemize}
    \item We propose spatio-temporal graph learning to address dynamic point cloud denoising. 
    The key idea is to exploit the inter-frame correlation of irregular point clouds for temporal consistency, which is inferred via spatial-temporal graph learning. 
    
    \item We pose spatial-temporal graph learning as the optimization of both spatial and temporal Mahalanobis distance metrics by minimizing Graph Laplacian Regularizer, where we decompose the positive-definite Mahalanobis distance metrics for computation efficiency.   
    
    \item Based on the spatial-temporal graph learning, we formulate dynamic point cloud denoising as joint optimization of the desired point cloud and underlying graph representation, and acquire the solution via alternating minimization.  
\end{itemize}

The outline of this paper is as follows. We first review previous static point cloud denoising methods in Section~\ref{sec:related}. 
Then we introduce basic concepts in spectral graph theory and GLR in Section~\ref{sec:graph}. 
Next, we elaborate on the proposed spatial-temporal graph learning in Section~\ref{sec:learning}, and present the proposed dynamic point cloud denoising algorithm based on spatial-temporal graph learning in Section~\ref{sec:method}.
Finally, experimental results and conclusions are presented in Section~\ref{sec:results} and Section~\ref{sec:conclude}, respectively.

\vspace{-0.05in}
\section{Related Work}
\label{sec:related}

To the best of our knowledge, there are few efforts on dynamic point cloud denoising in the literature. 
\cite{Schoenenberger2015Graph} provides a short discussion on how the proposed graph-based static point cloud denoising naturally generalizes to time-varying inputs such as 3D dynamic point clouds.  
Therefore, we discuss previous works on \textit{static} point cloud denoising, which can be divided into five classes: moving least squares (MLS)-based methods,
locally optimal projection (LOP)-based methods, sparsity-based methods, non-local methods, and graph-based methods.

\textbf{MLS-based methods.} MLS-based methods aim to approximate a smooth surface from the input point cloud and minimize the geometric error of the approximation. 
Alexa {\it et al.} approximate with a polynomial function on a local reference domain to best fit neighboring points in terms of MLS \cite{Alexa2003Computing}. 
Other similar solutions include algebraic point set surfaces (APSS) \cite{Guennebaud2007Algebraic} and robust implicit MLS (RIMLS) \cite{A2009Feature}. 
However, this class of methods may lead to over-smoothing results. 

\textbf{LOP-based methods.} LOP-based methods also employ surface approximation for denoising point clouds. 
Unlike MLS-based methods, the operator is non-parametric, which performs well in cases of ambiguous orientation. 
For instance, Lipman {\it et al.} define a set of points that represent the estimated surface by minimizing the sum of Euclidean distances to the data points \cite{Lipman2007Parameterization}. 
The two branches of \cite{Lipman2007Parameterization} are weighted LOP (WLOP) \cite{Hui2009Consolidation} and anisotropic WLOP (AWLOP) \cite{Huang2013Edge}. \cite{Hui2009Consolidation} produces a set of denoised, outlier-free and more evenly distributed particles over the original dense point cloud to keep the sample distance of neighboring points. 
\cite{Huang2013Edge} modifies WLOP with an anisotropic weighting function so as to preserve sharp features better. Nevertheless, LOP-based methods may also over-smooth point clouds.

\textbf{Sparsity-based methods.} These methods are based on sparse representation of point normals.
Regularized by sparsity, a global minimization problem is solved to obtain sparse reconstruction of point normals.
Then the positions of points are updated by solving another global $l_0$ \cite{Sun2015Denoising} or $l_1$ \cite{avron} minimization problem based on a local planar assumption.
Mattei {\it et al.} \cite{Mattei2017Point} propose \textit{Moving Robust Principal Components Analysis} (MRPCA) approach to denoise 3D point clouds via weighted $l_1$ minimization to preserve sharp features.
However, when locally high noise-to-signal ratios yield redundant features, these methods may not perform well and lead to over-smoothing or over-sharpening \cite{Sun2015Denoising}.

\textbf{Non-local methods.} Inspired by non-local means (NLM) \cite{buades2005non} and BM3D \cite{4271520} image denoising algorithms, this class of methods exploit self-similarities among surface patches in a point cloud. 
Digne {\it et al.} utilize a NLM algorithm to denoise static point clouds \cite{digne2012similarity}, while Rosman {\it et al.} deploy a BM3D method \cite{rosman2013patch}. 
Deschaud {\it et al.} extend non-local denoising (NLD) algorithm for point clouds, where the neighborhood of each point is described by the polynomial coefficients of the local MLS surface to compute point similarity \cite{deschaud10}.
\cite{sarkar2018structured} utilizes patch self-similarity and optimizes for a low rank (LR) dictionary representation of the extracted patches to smooth 3D patches.
Nevertheless, the computational complexity of these methods is usually high.

\textbf{Graph-based methods.}~~~This family of methods represent a point cloud on a graph, and design graph filters for denoising.
Schoenenberger {\it et al.} \cite{Schoenenberger2015Graph} construct a $k$-nearest-neighbor graph on the input point cloud and then formulate a convex optimization problem regularized by the gradient of the point cloud on the graph.
Dinesh {\it et al.} \cite{dinesh18arxiv} design a reweighted graph Laplacian regularizer for surface normals, which is deployed to formulate an optimization problem with a general lp-norm fidelity term that can explicitly model two types of independent noise.  
Zeng {\it et al.} \cite{zeng18arxiv} propose a low-dimensional manifold model (LDMM) with graph Laplacian regularization (GLR) and exploit self-similar surface patches for denoising. 
Instead of directly smoothing the 3D coordinates or surface normals, Duan {\it et al.} \cite{duan2018weighted} estimate the local tangent plane of each point based on a graph, and then reconstruct 3D point coordinates by averaging their projections on multiple tangent planes.
Hu {\it et al.} \cite{Hu19tsp} propose feature graph learning to optimize edge weights given available signal(s) assumed to be smooth with respect to the graph.  
However, the temporal dependency is not exploited yet in this class of methods.

\vspace{-0.05in}
\section{Background On Spectral Graph Theory}
\label{sec:graph}

We represent dynamic point clouds on undirected graphs. An undirected graph $\mathcal{G}=\{\mathcal{V},\mathcal{E},\mathbf{A}\}$ is composed of a node set $\mathcal{V}$ of cardinality $\left|\mathcal{V}\right|=n$, an edge set $\mathcal{E}$ connecting nodes, and a weighted adjacency matrix $\mathbf{A}$. $\mathbf{A} \in \mathbb{R}^{n \times n}$ is a real and symmetric matrix, where $a_{i,j} \geq 0$ is the weight assigned to the edge $(i,j)$ connecting nodes $i$ and $j$. Edge weights often measure the similarity between connected nodes. 

The graph Laplacian matrix is defined from the adjacency matrix. Among different variants of Laplacian matrices, the commonly used \textit{combinatorial graph Laplacian} \cite{shen10pcs,hu15tip} is defined as $ \mathbf{L}:=\mathbf{D}-\mathbf{A} $, where $ \mathbf{D} $ is the \textit{degree matrix}---a diagonal matrix where $ d_{i,i} = \sum_{j=1}^n a_{i,j} $.

Graph signal refers to data that resides on the nodes of a graph. In our case, the coordinates of each point in the input dynamic point cloud are the graph signal. A graph signal $ \vec{z} \in \mathbb{R}^n $ defined on a graph $ \mathcal{G} $ is smooth with respect to the topology of $ \mathcal{G} $ if  

\begin{equation}
	\sum_{i \sim j} a_{i,j}(z_i - z_j)^2 < \epsilon,
	\label{eq:prior}
\end{equation}
where $ \epsilon $ is a small positive scalar, and $ i \sim j $ denotes two nodes $i$ and $j$ are one-hop neighbors in the graph. In order to satisfy (\ref{eq:prior}), $ z_i $ and $ z_j $ have to be similar for a large edge weight $ a_{i,j} $, and could be quite different for a small $ a_{i,j} $. Hence, (\ref{eq:prior}) enforces $ \vec{z} $ to adapt to the topology of $ \mathcal{G} $, which is referred as  \textit{Graph Laplacian Regularizer} (GLR).

As $\vec{z}^T \mathbf{L} \vec{z} = \sum_{i \sim j} a_{i,j}(z_i - z_j)^2$ \cite{yale04lect2}, (\ref{eq:prior}) is concisely written as $ \vec{z}^T \mathbf{L} \vec{z} < \epsilon $ in the sequel. This term will be employed as the objective of spatio-temporal graph learning and the prior in the problem formulation of dynamic point cloud denoising.   




\vspace{-0.05in}
\section{Spatial-temporal Graph Learning}
\label{sec:learning}
\subsection{Formulation}
Spatial-temporal graph learning involves the inference of both edge connectivities and edge weights. 
In this paper, we focus on \textit{spatial-temporal edge weight learning} assuming the availability of connectivities in a spatial-temporal graph\footnote{We propose efficient construction of edge connectivities in Section~\ref{sec:method}. }. 
Since an edge weight $a_{i,j}$ for nodes $i$ and $j$ describes the pairwise similarity between the two nodes,  we essentially learn the {\it similarity distance metric}. 

While there exist various distance metrics, such as the Euclidean distance and bilateral distance \cite{tomasi1998bilateral}, we employ the \textit{Mahalanobis distance} \cite{mahalanobis1936} widely used in the machine learning literature, which takes feature correlations into account. 
Assuming a feature vector $\mathbf f_i \in \mathbb R^K$ is associated with each node $i$, the Mahalanobis distance between nodes $i$ and $j$ is defined as 
\begin{equation}
d_{\mathbf{M}}(\mathbf{f}_i,\mathbf{f}_j) = (\mathbf{f}_i - \mathbf{f}_j)^{\top} \mathbf{M} (\mathbf{f}_i - \mathbf{f}_j),
\label{eq:m_distance}
\end{equation}
where $\mathbf{M} \in \mathbb{R}^{K \times K} $ is the distance metric we aim to optimize. 
$\mathbf{M}$ is required to be positive definite, {\it i.e.}, $\mathbf{M} \succ 0$. 

Employing the commonly used Gaussian kernel \cite{Hu19tsp}, we formulate the edge weight as:
\begin{align}
    a_{i,j} &=\exp\left\{-d_{\mathbf{M}}(\mathbf{f}_i,\mathbf{f}_j)\right\} \\
    &= \exp\left\{-(\mathbf{f}_i-\mathbf{f}_j)^{\top} \mathbf{M} (\mathbf{f}_i-\mathbf{f}_j) \right\}.
\label{eq:weight}
\end{align}

We now pose an optimization problem for $\mathbf{M}$ with GLR (\ref{eq:prior}) as objective. 
we seek the optimal metric $\mathbf{M}$ that yields the smallest GLR term given feature vector $\mathbf{f}_i$ per node $i$ and point cloud observation with $n$ points $\{\mathbf v_i\}_{i=1}^n \in \mathbb R^{n \times 3}$, where $\mathbf v_i \in \mathbb R^3$ is the 3D coordinate vector of node $i$.
Specifically, denote the inter-node sample difference square of observation $\mathbf{z}$ in \eqref{eq:prior} by $d_{i,j} = \|\mathbf v_i-\mathbf v_j\|_2^2$ , we have 
\begin{equation}
\begin{split}
&\min_{\mathbf{M}}
    \sum_{i \sim j}\exp\left\{-(\mathbf{f}_i-\mathbf{f}_j)^{\top} \mathbf{M} (\mathbf{f}_i-\mathbf{f}_j) \right\} \, d_{i,j}\\
& \text{s.t.} \quad \, \mathbf{M} \succ 0.
\label{eq:optimize_L_new}
\end{split}
\end{equation}

Minimizing (\ref{eq:optimize_L_new}) directly would lead to one pathological solution, {\it i.e.}, $m_{i,i}=\infty, \forall i$, resulting in edge weight $a_{i,j} = 0$. 
As discussed in \cite{Hu19tsp}, this means nodes in the graph are all isolated, defeating the goal of finding a similarity graph. 
To avoid this solution, we constrain the trace of $\mathbf{M}$ to be smaller than a constant parameter $C>0$, resulting in
\begin{equation}
\begin{split}
&\min_{\mathbf{M}}
    \sum_{i \sim j} \exp\left\{-(\mathbf{f}_i-\mathbf{f}_j)^{\top} \mathbf{M} (\mathbf{f}_i-\mathbf{f}_j) \right\} \, d_{i,j}\\
& \text{s.t.} \quad \,\mathbf{M} \succ 0; \;\;\;
\mathrm{tr}(\mathbf{M}) \leq C.
\label{eq:optimize_c_constraint}
\end{split}
\end{equation}

It is nontrivial to solve $\mathbf{M}$ in low complexity as $\mathbf{M}$ is in the feasible space of a positive-definite cone. 
One na\"{i}ve approach is to employ gradient descent and then map the solution to the feasible space by setting negative eigenvalues of $\mathbf{M}$ to zero. 
Nevertheless, this would require eigen-decomposition of $\mathbf{M}$ per iteration with complexity $\mathcal O(n^3)$. 
Instead, since $\mathbf{M}$ is symmetric and positive definite, we propose to decompose $\mathbf{M}$ into  $\mathbf{M}=\mathbf{R}^{\top}\mathbf{R}$, where we consider $\mathbf{R} \in \mathbb{R}^{K \times K}$ assuming full rank\footnote{In general, $\mathbf{R}$ can be a non-square matrix.}. 
This leads to 
\begin{equation}
\begin{split}
d_{\mathbf{M}}(\mathbf{f}_i,\mathbf{f}_j)  
& =(\mathbf{f}_i - \mathbf{f}_j)^{\top} \mathbf{R}^{\top}\mathbf{R} (\mathbf{f}_i - \mathbf{f}_j) \\
& =(\mathbf{R} (\mathbf{f}_i - \mathbf{f}_j))^{\top}\mathbf{R} (\mathbf{f}_i - \mathbf{f}_j)\\
& = \|\mathbf{R} (\mathbf{f}_i - \mathbf{f}_j)\|_2^2,
\end{split}
\label{eq:m_distance}
\end{equation}
which can be viewed as a Euclidean distance of the linearly transformed feature distance via $\mathbf{R}$. 
The edge weight in \eqref{eq:weight} is then
\begin{equation}
    a_{i,j} = \exp\left\{-\|\mathbf{R} (\mathbf{f}_i - \mathbf{f}_j)\|_2^2 \right\}.
\label{eq:weight_R}
\end{equation}
Substituting \eqref{eq:m_distance} into \eqref{eq:optimize_c_constraint}, we have 
\begin{equation}
\begin{split}
&\min_{\mathbf{R}}
    \sum_{i \sim j}\exp\left\{-\|\mathbf{R} (\mathbf{f}_i - \mathbf{f}_j)\|_2^2 \right\} \, d_{i,j}\\
& \text{s.t.} \quad \,\mathrm{tr}(\mathbf{R}) \leq C; r_{i,i} \geq 0, \forall i.
\label{eq:optimize_linear}
\end{split} 
\end{equation}
Here we further constrain each diagonal entry of $\mathbf R$ to be non-negative. 
Otherwise, the diagonal entries might be negative with large absolute value that leads to infinite trace of $\mathbf M$.
Now the feasible space is converted to a convex set of a polytope, which can be solved much more efficiently. 

Further, we propose to learn {\it spatial} edge weights and {\it temporal} edge weights in spatio-temporal graph learning separately, which correspond to a spatial metric $\mathbf{R}_s$ and a temporal metric $\mathbf{R}_t$ respectively. 
Spatial edge weights are acquired by optimizing $\mathbf{R_s}$ from node pairs $\{i,j\}$ within the same frame, while temporal edge weights are computed from  $\mathbf{R_t}$ optimized via temporally connected node pairs $\{i,j\}$. 
Given appropriate node pairs, the optimization algorithms of $\mathbf{R_s}$ and $\mathbf{R_t}$ are the same, which will be discussed next. 


\subsection{Optimization Algorithm}



To solve the constrained optimization problem \eqref{eq:optimize_linear} efficiently, we employ a \textit{proximal gradient} (PG) approach \cite{Parikh31}.
We first define an indicator function $I_{\mathcal{S}}(\mathbf{R})$:
\begin{equation}
   I_{\mathcal{S}}(\mathbf{R})=
    \left\{
        \begin{array}{lr} 
            0, & \mathbf{R} \in \mathcal{S}  \\
            \infty, & \text{otherwise}
        \end{array}
    \right.
\end{equation}
where 
\begin{equation}
    \mathcal{S}=\left\{ \mathbf{R} ~\bigg|~ \mathrm{tr}(\mathbf{R}) \leq C; r_{i,i} \geq 0, \forall i \right\}.
\end{equation}

We then rewrite \eqref{eq:optimize_linear} as an unconstrained problem by incorporating the indicator function $I_{\mathcal{S}}(\mathbf{R})$ into the objective: 
 \begin{equation}
 \min_{\mathbf{R}}
     \sum_{i \sim j}\exp\left\{-\|\mathbf{R} (\mathbf{f}_i - \mathbf{f}_j)\|_2^2 \right\} \, d_{i,j} + I_{\mathcal{S}}(\mathbf{R}).
     \label{eq:solve_P}
 \end{equation}  
 The first term is convex with respect to $\mathbf{R}$ and differentiable, while the second term $I_{\mathcal{S}}(\mathbf{R})$ is convex but non-differentiable. 
 we can thus employ PG to solve (\ref{eq:solve_P}) as follows.
 
We first compute the gradient of the first term $F$ with respect to $\mathbf{R}$:
\begin{equation}
\begin{split}
    & \bigtriangledown F(\mathbf{R}) \\
    = &-2\sum_{\{i,j\}}\mathbf{R}(\mathbf{f}_i - \mathbf{f}_j) (\mathbf{f}_i - \mathbf{f}_j)^{\top}\exp\left\{-\|\mathbf{R} (\mathbf{f}_i - \mathbf{f}_j)\|_2^2 \right\} \, d_{i,j}.
\end{split}
\end{equation}

Next, we define a proximal mapping $\Pi_{I_{\mathcal{S}}}(\mathbf{V})$ for the second term---indicator function $I_{\mathcal{S}}(\mathbf{V})$---which is a projection onto the linear set $\mathcal{S}$, {\it i.e.},
 \begin{equation}
     \Pi_{I_{\mathcal{S}}}(\mathbf{V})=
     \left\{
         \begin{array}{lr} 
             \mathbf{V}, & \mathrm{tr}(\mathbf{R}) \leq C,r_{i,i} \geq 0, \forall i  \\
             g(\mathbf{V})-\alpha \cdot \text{diag}(g(\mathbf{V})), & \text{otherwise,} 
         \end{array}
     \right.
 \end{equation}
 where $g(\mathbf V)$ is a function that projects all diagonal entries of $\mathbf V$ to non-negative values, {\it i.e.}, $\max(r_{i,i},0), \forall i$. 
 $\alpha$ is any positive root of $\mathrm{tr}(g(\mathbf{V})-\alpha\cdot\text{diag}(g(\mathbf{V}))) = C$.

Each iteration of the PG algorithm can be now written as:
 \begin{equation}
     \mathbf{R}^{l+1}=\Pi_{I_{\mathcal{S}}}(\mathbf{R}^{l}-t\bigtriangledown F(\mathbf{R}^{l})),
     \label{eq:gradient_descent}
 \end{equation}
where $t$ is the step size, and $\mathbf{R}^{l}$ and $\mathbf{R}^{l+1}$ are the solved metric at the $l$-th and $(l+1)$-th iteration respectively. 
As discussed in \cite{Parikh31}, 
the algorithm will converge with rate $O(1/l)$ for a fixed step size $t^l = t \in (0,2/L]$, where $L$ is a Lipschitz constant that requires computation of the Hessian of $F$. 
In our experiment, we choose a small step size $t$ empirically, which is small enough to satisfy the Lipschitz smoothness of the objective function.
In the first iteration, we initialize $\mathbf{R}$ to be an identity matrix, and assign such a $C$ that $\mathrm{tr}(\mathbf{R}) \leq C$. 
We empirically find that the denoising results are relatively insensitive to $C$, as long as $C$ is slightly larger than the dimension $K$ of features per node. 


\vspace{-0.05in}
\section{Proposed Dynamic Point Cloud Denoising}
\label{sec:method}
\begin{figure*}
    \centering
    \includegraphics[width=0.8\linewidth]{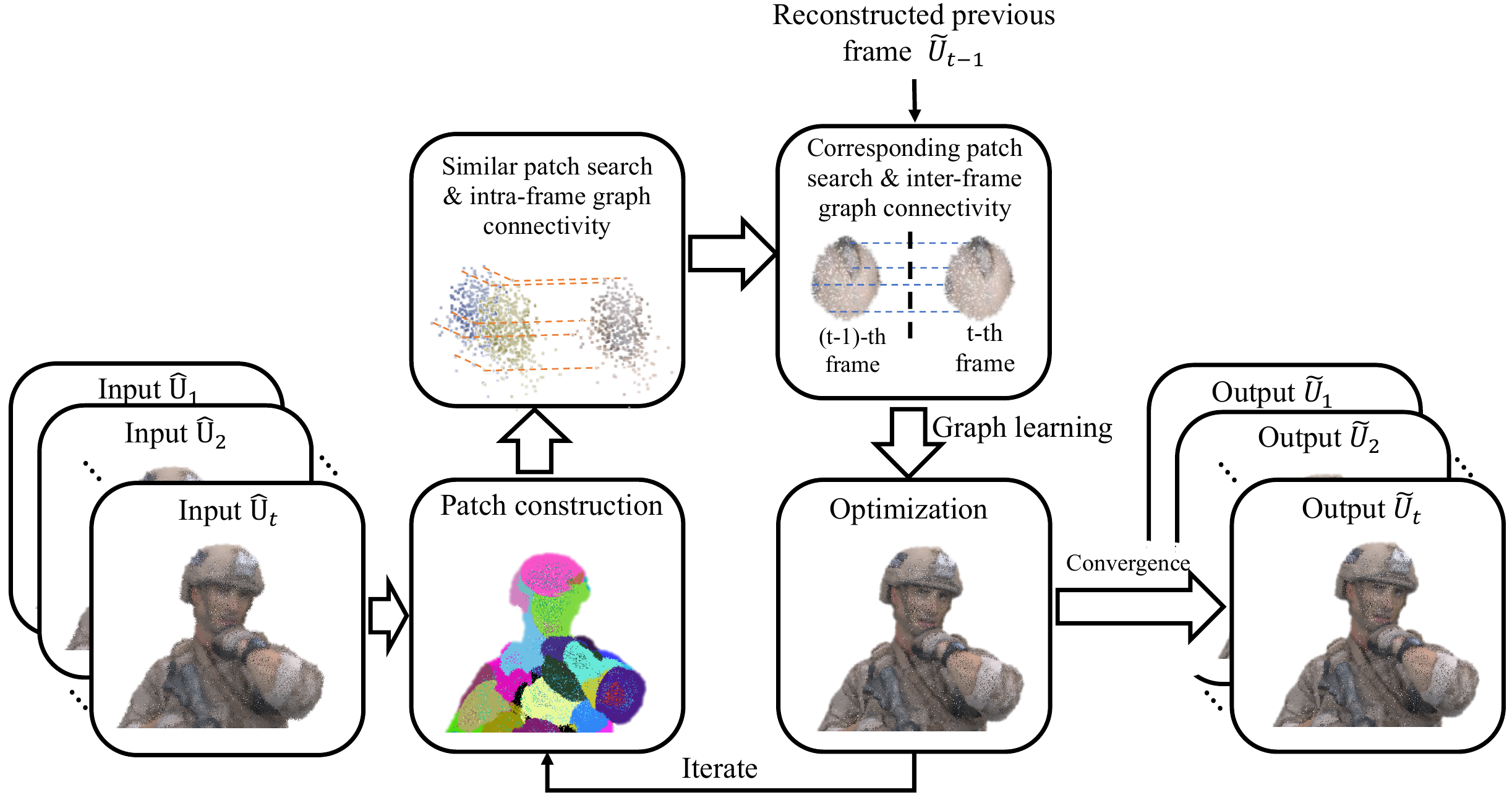}
    \centering
    \caption{The flowchart of the proposed dynamic point cloud denoising algorithm.}     
    \label{fig:flowchart}
\end{figure*}

Leveraging on the proposed spatio-temporal graph learning, we propose a dynamic point cloud denoising algorithm to exploit the spatio-temporal correlation. 
A dynamic point cloud sequence $\mathcal{P}=\{\mathbf{U}_1,\mathbf{U}_2,...,\mathbf{U}_m\}$ consists of $m$ frames of point clouds. The coordinates $\mathbf{U}_t=[\mathbf{u}_{t,1}, \mathbf{u}_{t,2}, ... , \mathbf{u}_{t,n}]^\top\in \mathbb{R}^{n\times3}$ denote the position of each point in the point cloud at frame $t$, in which $\mathbf{u}_{t,i} \in \mathbb{R}^3$ represents the coordinates of the $i$-th point in frame $t$. 
Let $\mathbf{U}_t$ denote the ground truth coordinates of the $t$-th frame, and $\hat{\mathbf{U}}_{t-1}$, $\hat{\mathbf{U}}_t$ denote the noise-corrupted coordinates of the $(t-1)$-th and $t$-th frame respectively.
Given each noisy frame $\hat{\mathbf{U}}_t$, we aim to recover its underlying signal $\mathbf{U}_t$, exploiting the intra-frame self-similarity in $\hat{\mathbf{U}}_t$ as well as the inter-frame dependencies with the {\it reconstructed} previous frame $\widetilde{\mathbf{U}}_{t-1}$.  

\subsection{Patch Construction}
In order to exploit local characteristics of point clouds, we model both intra-frame and inter-frame dependencies on \textit{patch} basis. Unlike images or videos defined on regular grids, point clouds reside on irregular domain with uncertain local neighborhood, thus the definition of a patch is nontrivial. We define a patch $\mathbf{p}_{t,l} \in \mathbb{R}^{(k+1) \times 3}$ in point cloud $\hat{\mathbf{U}}_t$ as a local point set of $k+1$ points, consisting of a centering point $\mathbf{c}_{t,l} \in \mathbb{R}^3$ and its $k$-nearest neighbors in terms of Euclidean distance. Then the entire set of patches at frame $t$ is     
\begin{equation}
    \mathbf{P}_t = \mathbf{S}_t \hat{\mathbf{U}}_t-\mathbf{C}_t,
    \label{eq:patch}
\end{equation}
where $\mathbf{S}_t\in\{0, 1\}^{(k+1)M\times n}$ is a sampling matrix to select points from point cloud $\hat{\mathbf{U}}_t$ so as to form $M$ patches of $(k+1)$ points per patch, and $\mathbf{C}_t \in \mathbb{R}^{(k+1)M\times 3}$ contains the coordinates of patch centers for each point.

Specifically, as each patch is formed around a patch center, we first select $M$ points from $\hat{\mathbf{U}}_t$ as the patch centers, denoted as $\{\mathbf{c}_{t, 1}, \mathbf{c}_{t, 2}, ... , \mathbf{c}_{t, M}\}\in \mathbb{R} ^ {M \times 3}\subset \hat{\mathbf{U}}_t$. In order to keep the patches distributed as uniformly as possible, we first choose a random point in $\hat{\mathbf{U}}_t$ as $\mathbf{c}_{t, 1}$, and add a point which holds the farthest distance to the previous patch centers as the next patch center, until there are $M$ points in the set of patch centers. We then search the $k$-nearest neighbors of each patch center in terms of Euclidean distance, which leads to $M$ patches in $\hat{\mathbf{U}}_t$.  

\subsection{Spatio-Temporal Similar Patch Search}
\label{section:Similar/Corresponding Patch Search}
Given each constructed patch in $\hat{\mathbf{U}}_t$, we search for its similar patches locally in $\hat{\mathbf{U}}_t$, and its corresponding patch in $\widetilde{\mathbf{U}}_{t-1}$ so as to exploit the spatio-temporal correlation. 
A metric is thus necessary to measure the similarity between patches, which remains challenging due to irregularity of patches. 
Taking two patches $\hat{\mathbf{p}}_{t,l}$ and $\hat{\mathbf{p}}_{t,m}$ in frame $\hat{\mathbf{U}}_t$ for instance, we discuss the similar patch search as follows. 
The temporally corresponding patches are searched in the same way. 

\subsubsection{Similarity Metric}
We deploy a simplified approach of \cite{zeng20183d} to measure the similarity between two patches $\hat{\mathbf{p}}_{t,l}$ and $\hat{\mathbf{p}}_{t,m}$. The key idea is to compare the distance from each point in the two patches to the tangent plane at one patch center, which captures the similarity in geometric curvature of patches.

Firstly, we structure the tangent planes of two patches. As a point cloud essentially characterizes the continuous surface of 3D objects, we calculate the surface normals $ \vec {n}_l $ and $\vec{n}_m $ for patch $\hat{\mathbf {p}}_ {t,l} $ and patch $ \hat{\mathbf{p}} _ {t, m } $ respectively for structural description. 
Then we acquire the tangent planes of the two patches at the patch center $\mathbf {c}_{t,l} $ and $ \mathbf {c}_{t,m} $ respectively.

Secondly, we measure patch similarity by the distance of each point in the two patches to the corresponding tangent plane. Specifically, we first project each point in patch $ \hat{\mathbf {p}} _ {t,l} $ and patch $ \hat{\mathbf{p}} _ {t,m} $ to the tangent plane of patch $ \hat{\mathbf {p}} _ {t , l } $. For the $i$-th point $\mathbf{v}_{l}^i$ in patch $ \hat{\mathbf {p}} _ {t , l } $, we search a point $\mathbf{v}_{m}^i$ in $\hat{\mathbf {p}} _ {t , m } $, whose projection on the tangent plane is closest to that of $\mathbf{v}_{l}^i$. Denote by $d_l(\mathbf{v}_l^i)$ and $d_l(\mathbf{v}_m^i)$ the distance of the two points to their projections on the tangent plane of patch $ \hat{\mathbf {p}} _ {t , l } $, $|d_l(\mathbf{v}_l^i) - d_l(\mathbf{v}_m^i)|$ is regarded as the difference of the two patches in point $\mathbf{v}_l^i$ and $\mathbf{v}_m^i$. Then we acquire the average difference between the two patches at all the $(k+1)$ points:
\begin{equation}
    {D}_{l,m} = \sqrt{\frac{1}{k+1}\sum_{i=1}^{k+1}\left[d_l(\mathbf{v}_l^i) - d_l(\mathbf{v}_m^i)\right]^2}.
\end{equation}
Similarly, projecting each point in patch $\hat{\mathbf{p}}_{t,l}$ and patch $\hat{\mathbf{p}}_{t,m}$ to the tangent plane of patch $\hat{\mathbf{p}}_{t,m}$, we acquire an average difference ${D}_{{m,l}}$.
The final mean difference between the two patches is
\begin{equation}
    \widetilde{{D}}_{l,m} = \sqrt{\frac{1}{2}({D}_{{l,m}}^2+{D}_{m,l}^2)}.
    \label{eq:patch_distance}
\end{equation}


\subsubsection{Local Patch Search} 
Given a target patch in the $t$-th frame $\hat{\mathbf{p}}_{t, l}, l\in \left[1, M\right]$, we seek its $r$ most similar patches within the current frame $\hat{\mathbf U}_t$ based on the similarity measure in \eqref{eq:patch_distance}, {\it i.e.}, $\{\hat{\mathbf{p}}_{t, m}\}_{m=1}^r$. 
As to the temporally corresponding patch in the previous frame $\widetilde{\mathbf{U}}_{t-1}$, we only search the most similar one to $\hat{\mathbf{p}}_{t, l}$ as the corresponding patch $\widetilde{\mathbf{p}}_{t-1, l}$. 

Further, in order to reduce the computation complexity of global search, we set a {\it local} window in the $t$-th frame for fast similar patch search, which contains patches centering at the $h$-nearest neighbors of the target patch center in terms of Euclidean distance. 
Then we evaluate the similarity between the target patch and these $h$-nearest patches instead of all the patches in the $t$-th frame, which significantly improves the search efficiency. 

Similarly, in the search of a temporally corresponding patch, we set a local window in the $(t-1)$-th frame $\widetilde{\mathbf{U}}_{t-1}$, which includes reference patches centering at the $h$-nearest neighbors of the collocated target patch center in terms of Euclidean distance. 
The reference patch with the largest similarity to the target patch is chosen as the temporally corresponding patch $\widetilde{\mathbf{p}}_{t-1, l}$. 


\vspace{-0.1in}

\begin{figure}[t]
    \centering
    \includegraphics[width=\linewidth]{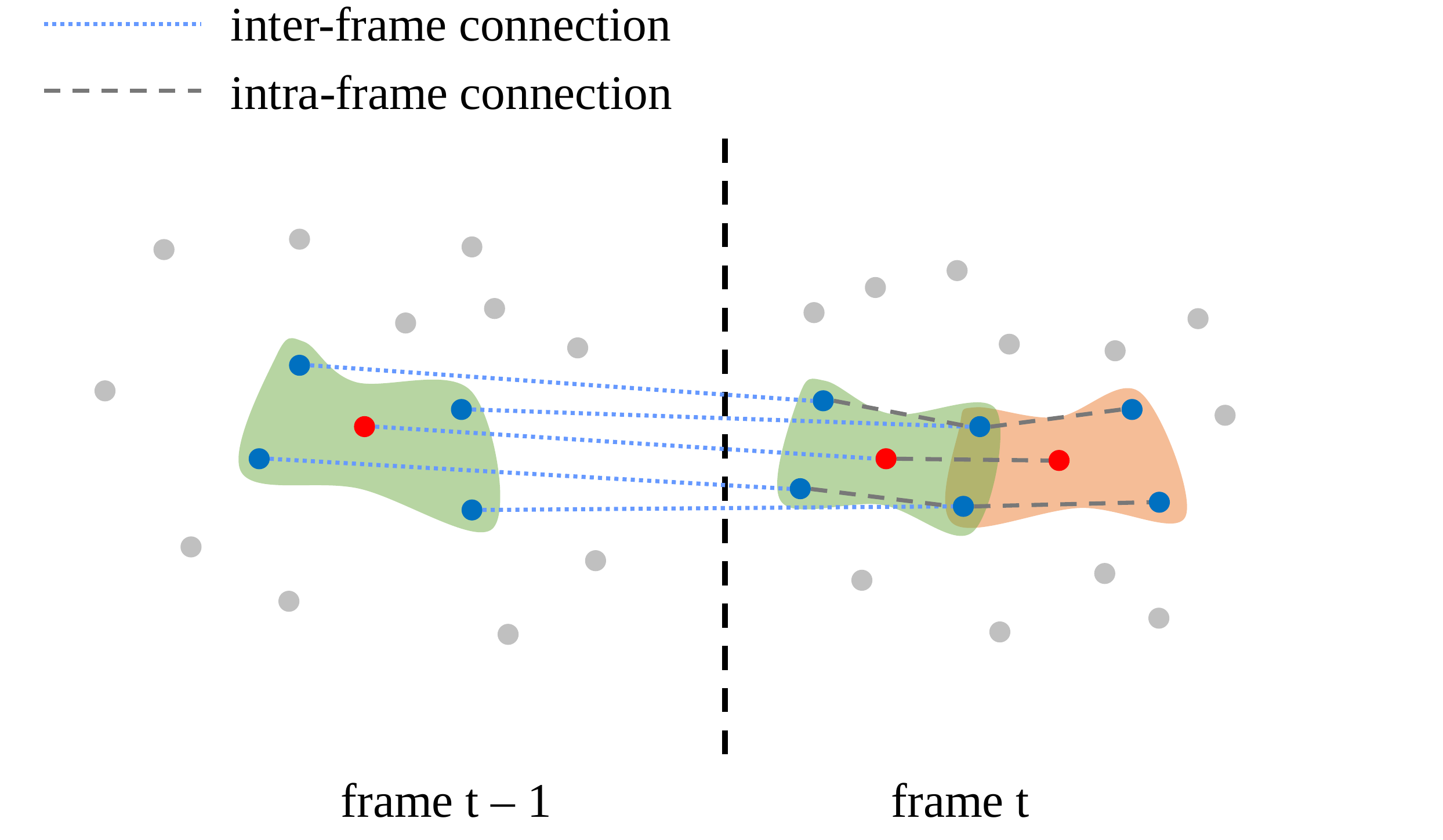}
    \vspace{-0.2in}
    \caption{Illustration of the spatio-temporal graph construction, including the inter-frame graph connection and intra-frame graph connection. In general, patches are overlapped within one frame, thus a point in one patch might be connected to its neighbor in the other patch in the overlapped region.}
    \label{fig:graph}
\end{figure}

\subsection{Proposed Spatio-Temporal Graph Connectivity}
\label{subsec:connecitivity}
Having searched intra-frame similar patches and inter-frame corresponding patches, we build a spatio-temporal graph over the patches. 
The construction of intra-frame edge connectivities and inter-frame edge connectivities is based on patch similarity, as demonstrated in Fig.~\ref{fig:graph}. 

\subsubsection{Intra-frame graph connectivities}
Given a target patch $\hat{\mathbf{p}}_{t, l}$ in the $t$-th frame $\hat{\mathbf U}_t$, we construct a graph between $\hat{\mathbf{p}}_{t, l}$ and each of its similar patches $\hat{\mathbf{p}}_{t, m}$. 
Specifically, we connect each point in $\hat{\mathbf{p}}_{t, l}$ with its nearest neighbor in $\hat{\mathbf{p}}_{t, m}$, where the distance is in terms of their projections onto the tangent plane with orientation orthogonal to the surface normal of $\hat{\mathbf{p}}_{t, l}$ at the patch center of $\hat{\mathbf{p}}_{t, l}$. 
Similarly, each point in $\hat{\mathbf{p}}_{t, m}$ is connected with the nearest point in $\hat{\mathbf{p}}_{t, l}$ in terms of their projections onto the tangent plane orthogonal to the surface normal of $\hat{\mathbf{p}}_{t, m}$ at the patch center of $\hat{\mathbf{p}}_{t, m}$. 
Note that, we do not connect points within each patch explicitly, though an edge may exist between two points in a patch if they are also nearest neighbors in two different patches due to patch overlaps, as shown in Fig.~\ref{fig:graph}. 

\subsubsection{Inter-frame graph connectivities} 
In order to leverage the inter-frame correlation and keep the temporal consistency, we connect corresponding patches between $\hat{\mathbf U}_t$ and $\widetilde{\mathbf U}_{t-1}$. 
Similarly to the intra-graph construction, we connect each point in patch $\hat{\mathbf{p}}_{t,l}$ with its nearest point in patch $\widetilde{\mathbf{p}}_{t-1, l}$, where the distance is in terms of their projections onto the tangent plane orthogonal to the surface normal of patch $\hat{\mathbf{p}}_{t,l}$ at the patch center of $\hat{\mathbf{p}}_{t,l}$. 
As such, we construct spatio-temporal graph connectivities as demonstrated in Fig.~\ref{fig:graph}. 

\vspace{-0.1in}

\subsection{Formulation of Dynamic Point Cloud Denoising}

We formulate dynamic point cloud denoising as an optimization problem for each underlying point cloud frame $\mathbf{U}_t$ and distance metrics $\{\mathbf{R}_s,\mathbf{R}_t\}$, taking into account both inter-frame consistency and intra-frame smoothness. 

Specifically, we seek the optimal $\mathbf{U}_t$, $\mathbf{R}_s$ and $\mathbf{R}_t$ to minimize an objective function including: 
1) a data fidelity term, which enforces $\mathbf{U}_t$ to be close to the observed noisy point cloud frame $\hat{\mathbf{U}}_t$; 
2) a temporal consistency term, which promotes the consistency between each patch $\mathbf{P}_{t,i} \in \mathbb{R}^{(k+1) \times 3}$ in $\mathbf{U}_t$ and its correspondence $\widetilde{\mathbf{P}}_{t-1,i}$ in the reconstructed previous frame $\mathbf{\widetilde{U}}_{t-1}$; 
3) a spatial smoothness term, which enforces smoothness of each patch in $\mathbf{U}_t$ with respect to the underlying graph encoded in the Laplacian $\mathbf{L}_{t}(\mathbf{R}_s) \in \mathbb{R}^{(k+1)M \times (k+1)M}$. 
The problem formulation is mathematically written as
\begin{equation}
\begin{split}
&\min\limits_{\mathbf{U}_t,\mathbf{R}_s,\mathbf{R}_t} ||\mathbf{U}_t-\hat{\mathbf{U}}_t||_2^2+\lambda_1\sum\limits_{i=1}^{M}{||\sqrt{\mathbf{w}_i(\mathbf{R}_t)^{\top}}(\mathbf{P}_{t,i}-\widetilde{\mathbf{P}}_{t-1,i})||_2^2}\\
&~~~~~~~~~~~+\lambda_2\text{tr}(\mathbf{P}_t^\top \mathbf{L}_{t}(\mathbf{R}_s)\mathbf{P}_t),\\
& \quad \text{s.t.} \quad \, \mathbf{P}_t = \mathbf{S}_t \mathbf{U}_t-\mathbf{C}_t,
\label{eq:final_2}
\end{split}
\end{equation}
where each element of $\mathbf{w}_i(\mathbf{R}_t) \in \mathbb{R}^{k+1}$ is the edge weight between a pair of temporally corresponding points for the $i$-th patch. $\lambda_1$ and $\lambda_2$ are weighting parameters for the trade-off among the data fidelity term, the temporal consistency term and the spatial smoothness term.
For the simplicity of notation, we define a diagonal matrix $\mathbf{W}_{t, t-1} \in \mathbb{R}^{(k+1)M \times (k+1)M}$ to describe the temporal weights $\mathbf{w}_i$ between corresponding patches:
\begin{equation}
\mathbf{W}_{t, t-1} = \text{diag}\left[\underbrace{\sqrt{w_{1,1}} ... \sqrt{w_{1,k+1}}}_{k+1} ... \underbrace{\sqrt{w_{M,1}} ... \sqrt{w_{M,k+1}}}_{k+1}\right].
\end{equation}

Substituting $\mathbf{W}_{t, t-1}$ and the constraint into the objective of \eqref{eq:final_2}, we have 
\begin{equation} 
\begin{split}
\min_{\mathbf{U}_t,\mathbf{R}_s,\mathbf{R}_t} \ \ \begin{split}&  \| \mathbf{U}_t - \hat{\mathbf{U}}_t \|_2^2  \\
& + \lambda_1 \| \mathbf{W}_{t,t-1}(\mathbf{R}_t)(\mathbf{S}_{t}\mathbf{U}_{t} - \mathbf{C}_{t}) - \mathbf{W}_{t,t-1}(\mathbf{R}_t) \widetilde{\mathbf{P}}_{t-1} \|_2^2 \\
& + \lambda_2 \text{tr}((\mathbf{S}_{t} \mathbf{U}_{t} - \mathbf{C}_{t})^\top \mathbf{L}_{t}(\mathbf{R}_s)(\mathbf{S}_{t} \mathbf{U}_{t} - \mathbf{C}_{t})).
\end{split}
\end{split}
\label{eq:final_3}
\end{equation} 
(\ref{eq:final_3}) is nontrivial to solve with three optimization variables. 
Next, we develop an alternating minimization approach to solve \eqref{eq:final_3}.

\subsection{Proposed Algorithm for Dynamic Point Cloud Denoising}
We propose to address \eqref{eq:final_3} by alternately optimizing the underlying point cloud frame $\mathbf{U}_t$ and the distance metrics $\mathbf{R}_s$ and $\mathbf{R}_t$. 
The iterations terminate when the difference in the objective between  two consecutive iterations stops decreasing.

In particular, we first perform denoising on the first frame of a point cloud sequence exploiting available intra-correlations ({\it i.e.}, $\lambda_1=0$). 
Then for the subsequent frame, we take advantage of the {\it reconstructed} previous frame as better reference than the noisy version. 

\subsubsection{Optimizing the distance metrics $\mathbf{R}_s$ and $\mathbf{R}_t$}
We first initialize $\mathbf{U}_t$ with the noisy observation $\hat{\mathbf{U}}_t$, and optimize $\mathbf{R}_s$ and $\mathbf{R}_t$ via spatio-temporal graph learning described in Section~\ref{sec:learning}. 
Spatial and temporal edge weights are then computed from the learned $\mathbf{R}_s$ and $\mathbf{R}_t$ respectively via \eqref{eq:weight_R}, which lead to $\mathbf{L}_t$ and $\mathbf{W}_{t, t-1}$ in \eqref{eq:final_3}. 

In particular, we consider two types of features on point clouds: Cartesian coordinates and angles between surface normals. 
We adopt surface normals to promote smoothness of the underlying surface. 
Specifically, we employ a function of the angle $\theta_{i,j}$ between two normals $\mathbf n_i$ and $\mathbf n_j$ as the feature difference in surface normals at points $i$ and $j$ as follows
\begin{equation}
    d_{\theta_{i,j}}=1-|\cos{\theta_{i,j}}|
    =1-|\frac{\mathbf n_i \cdot \mathbf n_j}{\|\mathbf n_i\|\|\mathbf n_j\|}|.
\end{equation}
Along with the feature difference in three-dimensional coordinates, we form a four-dimensional feature difference vector $\mathbf f_i-\mathbf f_j=[x_i-x_j,y_i-y_j,z_i-z_j,d_{\theta_{i,j}}]^\top$, where $x_i,y_i,z_i$ denote the coordinate of point $i$ and $x_j,y_j,z_j$ the coordinate of point $j$. 
Based on the constructed spatio-temporal graph connectives discussed in Section~\ref{subsec:connecitivity}, we then employ the feature difference at each pair of spatially connected points to learn the spatial metric $\mathbf R_s$, and the feature difference at each pair of temporally connected points to learn the temporal metric $\mathbf R_t$ as described in Section~\ref{sec:learning}. 

\subsubsection{Optimizing the point cloud $\mathbf{U}_t$}
With both $\mathbf{L}_t$ and $\mathbf{W}_{t, t-1}$ fixed from the learned $\mathbf{R}_s$ and $\mathbf{R}_t$, we take the derivative of (\ref{eq:final_3}) with respect to $\mathbf{U}_t$ and set the derivative to $0$. This leads to the closed-form solution of $\mathbf{U}_t$:
\begin{equation}
\begin{split}
& \left( \mathbf{I}+\lambda_1 \mathbf{S}_t^\top \mathbf{W}_{t,t-1}^\top \mathbf{W}_{t,t-1}\mathbf{S}_t+\lambda_2 \mathbf{S}_t^\top \mathbf{L}_t \mathbf{S}_t\right) \mathbf{U}_t \\
= \ & \hat{\mathbf{U}}_t+ 
\lambda_1\mathbf{S}_t^\top \mathbf{W}_{t,t-1}^\top \mathbf{W}_{t,t-1}(\mathbf{C}_t+\widetilde{\mathbf{P}}_{t-1}) + \lambda_2 \mathbf{S}_t^\top \mathbf{L}_t\mathbf{C}_t,
\label{eq:closed_form_U}
\end{split}
\end{equation} 
where $\mathbf{I} \in \mathbb{R}^{n \times n}$ is an identity matrix. 
\eqref{eq:closed_form_U} is a system of linear equations and thus can be solved efficiently.
Then we employ the acquired solution of $\mathbf{U}_t$ to update $\mathbf{R}_s$ and $\mathbf{R}_t$ in the subsequent iteration. 

A flowchart of the dynamic point cloud denoising algorithm is demonstrated in Fig.~\ref{fig:flowchart}, and an algorithmic summary is presented in Algorithm \ref{alg:Framwork}.

\begin{algorithm}[htb]
  \caption{3D Dynamic Point Cloud Denoising}  
  \SetKwInOut{Input}{~~Input}\SetKwInOut{Output}{Output}
  \label{alg:Framwork}
    \Input{A noisy dynamic point cloud sequence $\hat{\mathcal{P}}=\{\hat{\mathbf{U}}_1,\hat{\mathbf{U}}_2,...,\hat{\mathbf{U}}_m\}$} 
    
    \Output{Denoised dynamic point cloud sequence  $\widetilde{\mathcal{P}}=\{ \widetilde{\mathbf{U}}_1,\widetilde {\mathbf{U}}_2,...,\widetilde{\mathbf{U}}_m\}$} 
    \For{$\hat{\mathbf{U}}_t$ in $\hat{\mathcal{P}}$}{
        Initialize $\widetilde{\mathbf{U}}_t$ with $\hat{\mathbf{U}}_t$ \\
        Select $M$ points (set $\mathbf{C}_t$) as patch centers; \\
        \For{$\mathbf{c}_l$ in $\mathbf{C}_t$}{
            Find $k$-nearest neighbors of $\mathbf{c}_l$; \\
            Build patch $\hat{\mathbf{p}}_{t,l}$; \\
            Add $\hat{\mathbf{p}}_{t,l}$ to the patch set $\hat{\mathbf{P}}_t$; \\
        }
            \Repeat{convergence}{
            \For{$\hat{\mathbf{p}}_{t,l}$ in $\hat{\mathbf{P}}_t$}{
                Search the $r$ most similar patches of $\hat{\mathbf{p}}_{t,l}$ in $\hat{\mathbf U}_t$ in terms of the metric in \eqref{eq:patch_distance}; \\  
                \For {$\hat{\mathbf{p}}_{t, j}$ in $\hat{\mathbf{p}}_{t,l}$'s similar patches}{
                    Connect nearest points in $\hat{\mathbf{p}}_{t,l}$ and $\hat{\mathbf{p}}_{t, j}$; \\
                }
                Search the corresponding patch $\widetilde{\mathbf{p}}_{t-1,l}$ of $\hat{\mathbf{p}}_{t,l}$ in $\widetilde{U}_{t-1}$ via \eqref{eq:patch_distance}; \\
                Connect corresponding points in $\hat{\mathbf{p}}_{t,l}$ and $\widetilde{\mathbf{p}}_{t-1, l}$; \\
            }
            Learn the spatial metric $\mathbf R_s$ from spatially connected point pairs via \eqref{eq:optimize_linear}, and compute the intra-frame graph Laplacian $\mathbf{L}_t$ via \eqref{eq:weight_R}; \\
            Learn the temporal metric $\mathbf R_t$ from temporally connected point pairs via \eqref{eq:optimize_linear}, and compute the weight matrix $\mathbf{W}_{t,t-1}$ between corresponding patches via \eqref{eq:weight_R}; \\
            Solve \eqref{eq:closed_form_U} to update $\widetilde{\mathbf{U}}_t$; \\
        }
        $\widetilde{\mathbf{U}}_t$ serves as the input for denoising the next frame.
    }
\end{algorithm}  

\begin{figure}[H]
    \centering
    \includegraphics[width=\linewidth]{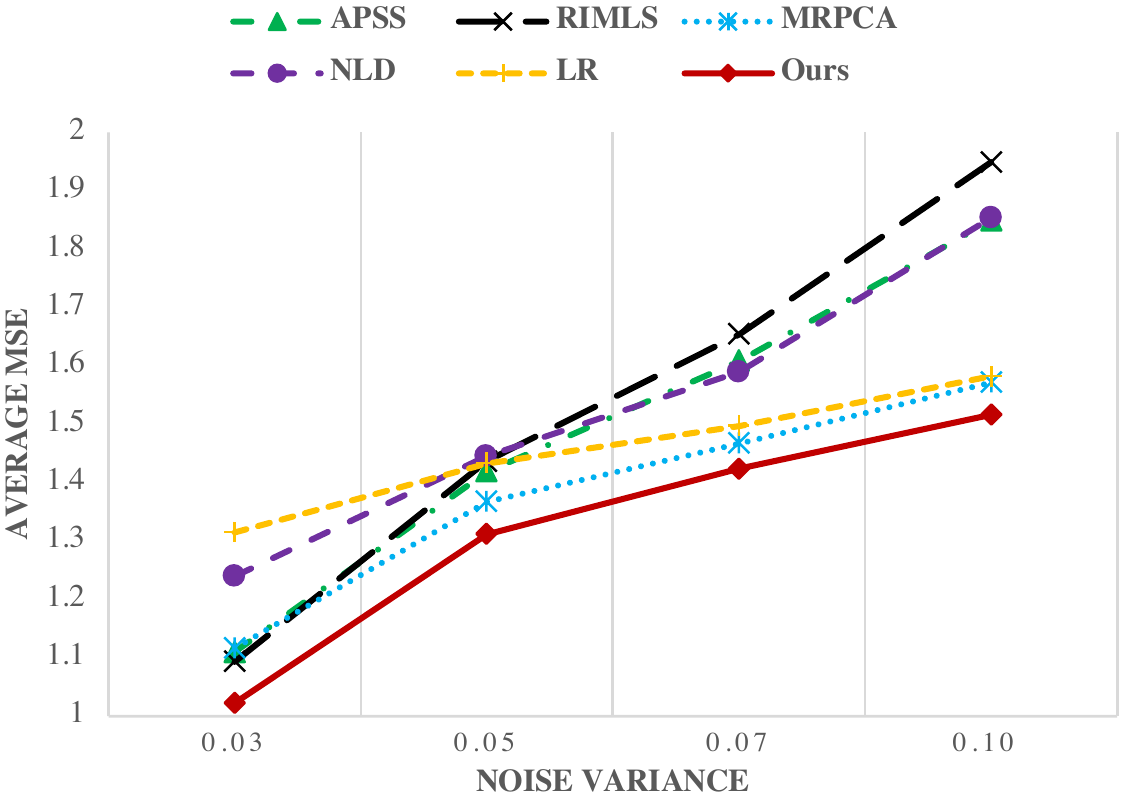}
    \vspace{-0.2in}
    \caption{Comparison between different methods in terms of the average MSE under various noise levels.}
    \label{fig:compare_chart}
    \vspace{-0.3cm}
\end{figure}


\vspace{-0.05in}
\section{Experimental Results}
\label{sec:results}
\begin{table*}[htbp]
\centering
\caption{MSE comparison for different models with Gaussian noise.}
\begin{tabularx}{0.9\textwidth}{|c|Y|Y|Y|Y|Y|Y|Y|Y|Y|}
\hline
\textbf{Model} & \textbf{Noisy} & \textbf{APSS} & \textbf{RIMLS} & \textbf{MRPCA} & \textbf{NLD} & \textbf{LR} & \textbf{Baseline1} & \textbf{Baseline2} & \textbf{Ours} \\ \hline \hline
\multicolumn{10}{|c|}{$\sigma=0.03$} \\ \hline
Soldier & 1.4984 & 1.4125 & 1.3572 & 1.3488 & 1.6148 & 1.7850 & 1.2979 & 1.3227  & $\mathbf{1.2976}$ \\ \hline
Longdress & 1.4746 & 1.3985 & 1.3360 & 1.3247 & 1.5900 & 1.7751 & 1.2722  & 1.2996  & $\mathbf{1.2704 }$ \\ \hline
Loot & 1.4715 & 1.3571 & 1.3279 & 1.3101 & 1.5751 & 1.6997 & 1.2434 & 1.2618 & $\mathbf{1.2433}$ \\ \hline
Redandblack & 1.4589 & 1.3892 & 1.3499 & 1.3221 & 1.6040 & 1.7701 & 1.2663 & 1.3024 & $\mathbf{1.2663}$ \\ \hline
Andrew & 0.8874 & 0.8839 & 0.8894 & 0.9655 & 0.9582 & 0.9745 & 0.8294 & 0.8302 & $\mathbf{0.8283}$\\ \hline
David & 0.9421 & 0.9131 & 0.9247 & 0.9487 & 0.9581 & 0.9558 & 0.8516 & 0.8527 & $\mathbf{0.8513}$ \\ \hline
Phil & 0.8732 & 0.8686 & 0.8752 & 0.9553 & 0.9589 & 0.9706 & 0.8171 & 0.8163 & $\mathbf{0.8154}$ \\ \hline
Ricardo & 0.9113 & 0.8994 & 0.9042 & 0.9449 & 0.9567 & 0.9587 & 0.8251  & 0.8275 & $\mathbf{0.8250}$  \\ \hline
Sarah & 0.8808 & 0.8677 & 0.8731 & 0.9257 & 0.9394 & 0.9425 & $0.8045$ & 0.8079  & $\mathbf{0.8045}$ \\ \hline 
Average & 1.1554 & 1.1100 & 1.0931 & 1.1162 & 1.2395 & 1.3147 & 1.0231 & 1.0357 & $\mathbf{1.0225}$ \\ \hline 
\hline
\multicolumn{10}{|c|}{$\sigma=0.05$} \\ \hline
Soldier & 2.1453 & 1.8047 & 1.8105 & 1.8116 & 1.9465 & 1.9767 & 1.7654 & 1.7907 & $\mathbf{1.7606}$ \\ \hline
Longdress & 2.1260 & 1.8007 & 1.7955 & 1.7922 & 1.9306 & 1.9601 & 1.7428  & 1.7839  & $\mathbf{1.7382 }$ \\ \hline
Loot & 2.1286 & 1.7668 & 1.7883 & 1.7703 & 1.9317 & 1.8949 & 1.7121  & 1.7320  & $\mathbf{1.7018}$ \\ \hline
Redandblack & 2.1110 & 1.7915 & 1.8061 & 1.7939 & 1.9389 & 1.9650 & 1.7423  & 1.7871  & $\mathbf{1.7351}$ \\ \hline
Andrew & 1.1516 & 1.1154 & 1.1359 & 1.0433 & 1.0532 & 1.0331 & 0.9852 & 0.9859 & $\mathbf{0.9841}$ \\ \hline
David & 1.2014 & 1.1562 & 1.1817 & 1.0217 & 1.0598 & 1.0123 & 0.9703 & 0.9713 & $\mathbf{0.9701}$\\ \hline
Phil & 1.1478 & 1.1027 & 1.1253 & 1.0426 & 1.0541 & 1.0351 & 0.9864 & 0.9857 & $\mathbf{0.9844}$ \\ \hline
Ricardo & 1.1720 & 1.1276 & 1.1541 & 1.0220 & 1.0513 & 1.0160 & 0.9881 & 0.9602  & $\mathbf{0.9596}$ \\ \hline
Sarah & 1.1533 & 1.1020 & 1.1281 & 1.0107 & 1.0376 & 1.0035 & 0.9705 & 0.9701 & $\mathbf{0.9686}$ \\ \hline 
Average & 1.5930 & 1.4186 & 1.4362 & 1.3676 & 1.4449 & 1.4330 & 1.3181 & 1.3297 & $\mathbf{1.3114}$ \\ \hline 
\hline
\multicolumn{10}{|c|}{$\sigma=0.07$} \\ \hline
Soldier & 2.5417 & 1.9675 & 2.0450 & 1.9999 & 2.156 & 2.0758 & 1.9551 & 1.9753 & $\mathbf{1.9356}$ \\ \hline
Longdress & 2.5139 & 1.9630 & 2.0297 & 1.9754 & 2.1324 & 2.0561 & 1.9409 & 1.9667 & $\mathbf{1.9275}$ \\ \hline
Loot & 2.5205 & 1.9359 & 2.0271 & 1.9487 & 2.1276 & 1.9950 & 1.8951  & 1.8990  & $\mathbf{1.8848}$ \\ \hline
Redandblack & 2.5035 & 1.9726 & 2.0537 & 1.9849 & 2.1421 & 2.0639 & 1.9807  & 1.9807  & $\mathbf{1.9284}$ \\ \hline
Andrew & 1.3241 & 1.3113 & 1.3295 & 1.0739 & 1.1482 & 1.0681 & 1.0408 & 1.0390 & $\mathbf{1.0378}$ \\ \hline
David & 1.3903 & 1.3821 & 1.3988 & 1.0516 & 1.1690 & 1.0519 & 1.0263 & 1.0262 & $\mathbf{1.0256}$ \\ \hline
Phil & 1.3202 & 1.2984 & 1.3192 & 1.0774 & 1.1423 & 1.0739 & 1.0456 & 1.0445 & $\mathbf{1.0431}$ \\ \hline
Ricardo & 1.3524 & 1.3388 & 1.3580 & 1.0528 & 1.1542 & 1.0542 & 1.0116 & 1.0121  & $\mathbf{1.0103}$ \\ \hline
Sarah & 1.3292 & 1.3080 & 1.3291 & 1.0413 & 1.1293 & 1.0390 & 1.0211 & 1.0230 & $\mathbf{1.0208}$ \\ \hline 
Average & 1.8662 & 1.6086 & 1.6545 &1.4673 &1.5890 &1.4975 &1.4352 &1.4407 & $\mathbf{1.4238}$  \\ \hline 
\hline
\multicolumn{10}{|c|}{$\sigma=0.10$} \\ \hline
Soldier & 3.0127 & 2.1404 & 2.3901 & 2.1874 & 2.4887 & 2.1813 & 2.1097  & 2.1049 & $\mathbf{2.0610}$ \\ \hline
Longdress & 2.9761 & 2.1236 & 2.3748 & 2.1360 & 2.4683 & 2.1506 & 2.0650  & 2.0941  & $\mathbf{2.0475}$ \\ \hline
Loot & 2.9853 & 2.1118 & 2.3338 & 2.1037 & 2.4664 & 2.0935 & 2.0080  & 2.0224 & $\mathbf{1.9941}$ \\ \hline
Redandblack & 2.9622 & 2.1433 & 2.3266 & 2.1563 & 2.4636 & 2.1640 & 2.0814  & 2.1213 & $\mathbf{2.0586}$ \\ \hline
Andrew & 1.5615 & 1.5916 & 1.5911 & 1.1203 & 1.3473 & 1.1326 & 1.1149 & 1.1098 & $\mathbf{1.1087}$ \\ \hline
David & 1.6621 & 1.7045 & 1.6959 & 1.1166 & 1.4193 & 1.1400 & 1.0772  & 1.0710  & $\mathbf{1.0682}$  \\ \hline
Phil & 1.5581 & 1.5850 & 1.5849 & 1.1296 & 1.3288 & 1.1408 & 1.1333 & 1.1257 & $\mathbf{1.1246}$ \\ \hline
Ricardo & 1.6068 & 1.6392 & 1.6390 & 1.1098 & 1.3698 & 1.1333 & 1.0811 & 1.0800 & $\mathbf{1.0800}$ \\ \hline
Sarah & 1.5740 & 1.6000 & 1.6031 & $\mathbf{1.0829}$ & 1.3284 & 1.1044  & 1.1078 & 1.1041 & 1.1035  \\ \hline
Average & 2.2110 & 1.8488 & 1.9488 & 1.5714 & 1.8534 & 1.5823 & 1.5309 & 1.5370 & $\mathbf{1.5162}$  \\ \hline 
\end{tabularx}
\label{tb:mse}

 \end{table*}
 
 \begin{table*}[htbp]
\centering
\caption{SNR comparison for different models with Gaussian noise.}
\begin{tabularx}{0.9\textwidth}{|c|Y|Y|Y|Y|Y|Y|Y|Y|Y|}
\hline
\textbf{Model} & \textbf{Noisy} & \textbf{APSS} & \textbf{RIMLS} & \textbf{MRPCA} & \textbf{NLD} & \textbf{LR} & \textbf{Baseline1} & \textbf{Baseline2} & \textbf{Ours} \\ \hline \hline
\multicolumn{10}{|c|}{$\sigma=0.03$} \\ \hline
Soldier & 60.5080  &  61.0895 &	61.4889 &	58.8487   & 59.7501 &	58.7476 &	61.9348 &	61.7464 &	$\mathbf{61.9364}$  \\ \hline
Longdress &  60.2722  & 61.0876 &	61.3056 &	58.9989    & 59.7318 &	58.6296 &	61.9514 &	61.7425 &	$\mathbf{61.9765}$  \\ \hline
Loot & 60.4739   & 61.0155 &	61.4740 &	58.7771    & 59.5969 &	58.8364 &	61.9617 &	61.8158 &	$\mathbf{61.9625}$  \\ \hline
Redandblack & 61.9718  &  62.4682 &	62.7552 &	60.0590   & 61.0298 &	60.0439 &	63.3921 &	63.1114 &	$\mathbf{63.3925}$ \\ \hline
Andrew & 60.6087   &  60.6634 &	60.6013 &	59.7802   & 59.8401 &	59.6678 &	61.3751 &	61.3089 &	$\mathbf{61.3751}$ \\ \hline
David & 61.8615   &  62.1288 &	62.0025 &	61.7464   & 61.6924 &	61.7124 &	62.9836 &	62.9715 &	$\mathbf{62.9789}$ \\ \hline
Phil & 61.0185   & 61.0715 &	60.9958 &	60.1201   & 60.0825 &	59.9568 &	61.8027 &	61.7344 &	$\mathbf{61.8022}$  \\ \hline
Ricardo & 61.8591  & 61.3581 &	61.3049 &	60.8649   & 61.3659 &	61.3384 &	62.8508 &	62.8199 &	$\mathbf{62.8502} $  \\ \hline
Sarah & 61.5861   &  61.7562 &	61.6941 &	61.1092   &60.9420 &	60.9047 &	62.4779 &	62.4360 &	$\mathbf{62.4775}$  \\ \hline 
Average &  61.1289 	&61.4043 &	61.5136 &	60.0338 &	60.4480 &	59.9820 &	62.3033 &	62.1874 &	$\mathbf{62.3057}$    \\ \hline 
\hline
\multicolumn{10}{|c|}{$\sigma=0.05$} \\ \hline
Soldier & 56.9009   &  58.6395 &	58.6073 &	57.2701   & 57.8826 &	57.7279 &	58.8591 &	58.7170 &	$\mathbf{58.9019} $ \\ \hline
Longdress & 56.6003   & 58.4490 &	58.3288 &	57.2296   & 57.7906 &	57.6380 &	58.8139 &	58.5815 &	$\mathbf{58.8409 }$ \\ \hline
Loot & 56.8114  & 58.4876 &	58.5077 &	57.1579  & 57.6307 &	57.7497 &	58.7638 &	58.6480 &	$\mathbf{58.8241}$ \\ \hline
Redandblack & 58.2667  &  59.9247 &	59.8431 &	58.4732  &  59.1330 &	58.9989 &	60.2008 &	59.9473 &	$\mathbf{60.2422} $  \\ \hline
Andrew & 58.0028  &  58.3371 &	58.1550 &	59.0052  &  58.8956 &	59.0832 &	59.6320 &	59.5540 &	$\mathbf{59.6384}$  \\ \hline
David & 59.4297  &  59.7683 &	59.5502 &	61.0052   & 60.6839 	&61.1377 &	61.6321 &	61.6308 &	$\mathbf{61.6428}$ \\ \hline
Phil & 58.2839  & 58.6850 &	58.4821 &	59.2455  & 59.1356 &	59.3136 &	59.8755 &	59.7854 &	$\mathbf{59.8823}$  \\ \hline
Ricardo & 59.3394  &  59.0967 &	58.8644 &	60.0804   & 60.4224 &	60.7580 &	61.3232 &	61.2826 &	$\mathbf{61.3291} $ \\ \hline
Sarah & 58.8916  &  59.3658 &	59.1317 &	60.2308  & 59.9485 &	60.2776 &	60.7052 &	60.6659 &	$\mathbf{60.7102} $ \\ \hline 
Average & 58.0585 &	58.9726 &	58.8300 &	58.8553 &	59.0581 &	59.1872 &	59.9784 &	59.8681 &	$\mathbf{60.0013}$  \\ \hline 
\hline
\multicolumn{10}{|c|}{$\sigma=0.07$} \\ \hline
Soldier &  55.2291  & 57.7755 &	57.3897 &	56.7700  & 56.8600 &	57.2386 &	57.8381 &	57.7361 &	$\mathbf{57.9548} $ \\ \hline
Longdress & 54.9225   & 57.5355 &	57.0761 &	56.7004  & 56.7966 &	57.1599 &	57.8408 &	57.6060 &	$\mathbf{57.8578} $ \\ \hline
Loot & 55.1564   &57.6245 &	57.2908 &	56.6518  & 56.5903 &	57.2345 &	57.8537 &	57.7271 &	$\mathbf{57.8798 }$ \\ \hline
Redandblack & 56.5859  &  58.9614 &	58.5588 &	57.9666   & 58.1366 &	58.5079 &	59.1883 &	58.9168 &	$\mathbf{59.2297 }$ \\ \hline
Andrew & 56.6060   & 56.7192 &	56.5814 &	58.7164   & 58.0317 	&58.7508 &	59.0601 &	59.0068 &	$\mathbf{59.0885 }$ \\ \hline
David & 57.9697  & 57.9839 &	57.8638 &	60.7169  & 59.7034 &	60.7551 &	61.1491 &	61.1794 &	$\mathbf{61.1718 }$ \\ \hline
Phil &  56.8851  & 57.0514 &	56.8925 &	58.9173  & 58.3323 &	58.9459 &	59.2335 &	59.1642 &	$\mathbf{59.2661 }$ \\ \hline
Ricardo & 57.9082  & 57.3803 &	57.2380 &	59.7839  & 59.4909 &	60.3900 &	60.8283 &	60.8178 &	$\mathbf{60.8535 }$ \\ \hline
Sarah &  57.4719  &  57.6519 &	57.4919 &	59.9323   & 59.1013 &	59.9301 &	60.1309 &	60.1041 &	$\mathbf{60.1467 }$ \\ \hline 
Average & 56.5261 &	57.6315 &	57.3759 &	58.4617 &	58.1159 &	58.7681 &	59.2359 &	59.1398 &	\textbf{59.2721} \\ \hline 
\hline
\multicolumn{10}{|c|}{$\sigma=0.10$} \\ \hline
Soldier & 53.5216 & 56.9332 &	55.8305 &	56.4273  & 55.7253 &	56.7428 &	57.0740 &	57.0984 &	$\mathbf{57.3077 }$ \\ \hline
Longdress & 53.1795 & 56.6672 &	55.6671 &	56.3759  & 55.6710 &	56.7106 &	57.1137 &	56.9744 &	$\mathbf{57.1986 }$ \\ \hline
Loot & 53.4494 & 56.8385 &	55.7228 &	56.3343 & 55.4772 &	56.7527 &	57.1631 &	57.0917 &	$\mathbf{57.2321 }$ \\ \hline
Redandblack & 54.8903 & 58.1316 &	57.0985 &	57.6910  & 57.0420 &	58.0349 &	58.4176 &	58.2291 &	$\mathbf{58.5282 }$ \\ \hline
Andrew & 54.9534 & 54.7820 &	54.7851 &	58.2932  & 56.4331 &	58.1644 &	58.1091 &	58.1559 &	$\mathbf{58.3518 }$ \\ \hline
David & 56.1841 & 55.8875 &	55.9380 &	60.1174  & 57.7633 &	59.9503 &	60.4625 &	60.5199 &	$\mathbf{60.5469 }$  \\ \hline
Phil & 55.2281 & 55.0572 &	55.0578 &	58.4442  & 56.8196 &	58.3415 &	58.1000 &	58.1810 &	$\mathbf{58.3842 }$ \\ \hline
Ricardo & 56.1872 & 55.3559 &	55.3571 &	59.2565  & 57.7833 &	59.6694 &	60.1116 &	60.0769 &	$\mathbf{60.2444 }$ \\ \hline
Sarah & 55.7812 & 55.6373 &	55.6179 &	\textbf{59.5410}  & 57.4776 &	59.3387 &	59.1646 &	59.2422 &	59.3179 \\ \hline
Average & 54.8194 &	56.1434 &	55.6750 &	{58.0534} &	56.6880 &	58.1895 &	58.4129 &	58.3966 &	$\textbf{58.5680}$ \\ \hline 
\end{tabularx}
\label{tb:snr}
 \end{table*}

\vspace{-0.05in}
\subsection{Experimental Setup}
\label{subsec:setup}
We evaluate our algorithm by testing on two benchmarks, including four MPEG sequences (\textit{Longdress}, \textit{Loot}, \textit{Redandblack} and \textit{Soldier}) from \cite{MPEG} and five MSR sequences (\textit{Andrew}, \textit{David}, \textit{Phil}, \textit{Ricardo} and \textit{Sarah}) from \cite{loop2016microsoft}.
We randomly choose six consecutive frames as the sample data: frame $601$-$606$ in \textit{Soldier}, frame $1201$-$1206$ in \textit{Loot}, frame $1201$-$1206$ in \textit{Longdress}, frame $1501$-$1506$ in \textit{Redandblack}, frame $61$-$66$ in \textit{Andrew}, frame $61$-$66$ in \textit{David}, frame $61$-$66$ in \textit{Phil}, frame $61$-$66$ in \textit{Ricardo} and frame $61$-$66$ in \textit{Sarah}.
We perform down-sampling with the sampling rate of $0.05$ prior to denoising since the number of points in each frame is about $1$ million. 
Because point clouds in the datasets are clean in general, we add white Gaussian noise with a range of variance $\sigma=\{0.03, 0.05, 0.07, 0.1\}$. 

Due to the lack of previous dynamic point cloud denoising approaches, we compare our algorithm with five competitive static point cloud denoising methods: APSS \cite{Guennebaud2007Algebraic}, RIMLS \cite{A2009Feature}, MRPCA \cite{Mattei2017Point}, NLD \cite{deschaud10}, and LR \cite{sarkar2018structured}. 
We perform each static denoising method frame by frame independently on dynamic point clouds. 
We employ two evaluation metrics: Mean Squared Error (MSE) and Signal-to-Noise Ratio (SNR) as in \cite{duan2018weighted}.

The parameter settings are as follows. 
We assign the upperbound $C$ of the trace of $\mathbf R$ in \eqref{eq:optimize_linear} as $10$.
The step size $t$ of the PG algorithm in \eqref{eq:gradient_descent} is assigned as $0.0001$. 
In \eqref{eq:final_3}, the weighting parameter of the temporal consistency term $\lambda_1$ is set to $0.003$, 
while that of the spatial smoothness term $\lambda_2$ is set to $0.1$.
Besides, given the first frame in each dataset, we set $\lambda_1=0$ as there is no previous frame for reference.
We divide each point cloud into $M=0.2n$ patches, where $n$ is the number of points in the point cloud. Each patch is connected with $r=8$ most similar patches spatially.

\vspace{-0.05in}
\subsection{Experimental Results}
\subsubsection{Objective results} 
We list the denoising results of comparison methods measured in MSE and SNR respectively in Tab.~\ref{tb:mse} and Tab.~\ref{tb:snr}, and mark the lowest MSE/SNR in bold. 
Our method outperforms all the five static point cloud denoising approaches on the nine datasets at all the noise levels on average. 
Specifically, we achieve reduction of the MSE by $11.90\%$ on average over APSS, $14.00\%$ on average over RIMLS, $4.50\%$ on average over MRPCA, $13.92\%$ on average over NLD, and $9.50\%$ on average over LR. 
The results are also visualized in Fig.~{\ref{fig:compare_chart}} for easy comparison. 
This validates the effectiveness of our method. 


\subsubsection{Ablation studies}
To evaluate the two main contributions of our method---temporal consistency and spatio-temporal graph learning, we conduct two baselines for comparison. 
In \textit{Baseline1}, we remove the temporal consistency term in \eqref{eq:final_3} to evaluate the importance of temporal references, {\it i.e.}, $\lambda_1=0$. 
Only similar patches in the same frame are employed as the reference for denoising in this case, where the underlying spatial graph is acquired from the proposed learning of spatial distance metric. 
For the evaluation of spatio-temporal graph learning, in \textit{Baseline2} we set edge weights as a Gaussian function with manually assigned parameters $\mathbf M=\mathbf I$ instead of the proposed graph learning, {\it i.e.}, $ a_{i,j} = \exp\left\{-(\mathbf{f}_i-\mathbf{f}_j)^{\top} (\mathbf{f}_i-\mathbf{f}_j) \right\}$.

As presented in Tab.~\ref{tb:mse} and Tab.~\ref{tb:snr}, we outperform both \textit{Baseline1} and \textit{Baseline2} constantly.  
Specifically, we achieve reduction of MSE over \textit{Baseline1} by $0.63\%$ on average at all noise levels, and over \textit{Baseline2} by $1.30\%$ on average. 
In particular, we achieve larger gain over dynamic point clouds with {\it slower motion} due to the stronger temporal correlation. 
For instance, the average MSE reduction over the comparatively static \textit{Redandblack} is $1.16\%$, while that over \textit{Sarah} is $0.17\%$ with much more dynamic motion in the tested frames. 
Further, the MSE reduction over \textit{Baseline1} is $0.06\%,0.51\%,0.79\%,0.96\%$ respectively with increasing noise levels. 
This indicates that the temporal correlation plays a more important role at high noise levels. 


\subsubsection{Subjective results} 
\begin{figure*}[htbp]
    \centering
    \subfigure[Ground-truth]{\includegraphics[width=\textwidth]{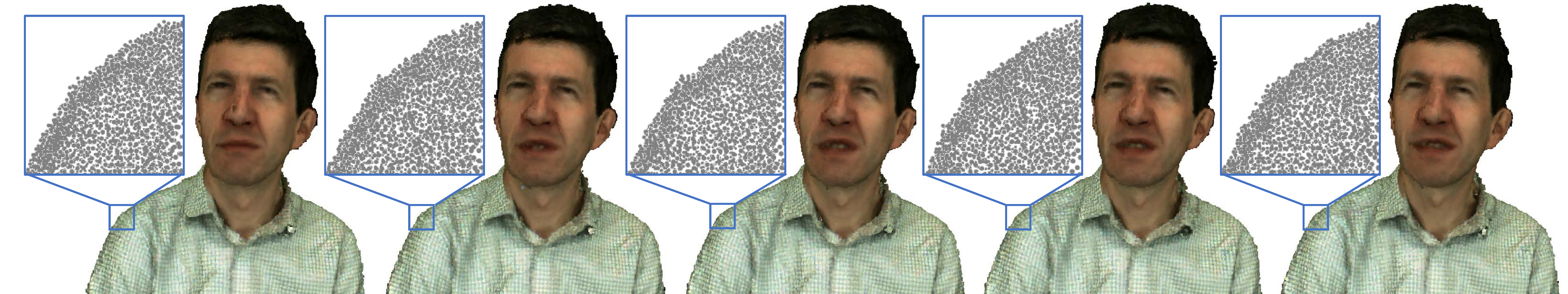}}
    \subfigure[Noisy]{\includegraphics[width=\textwidth]{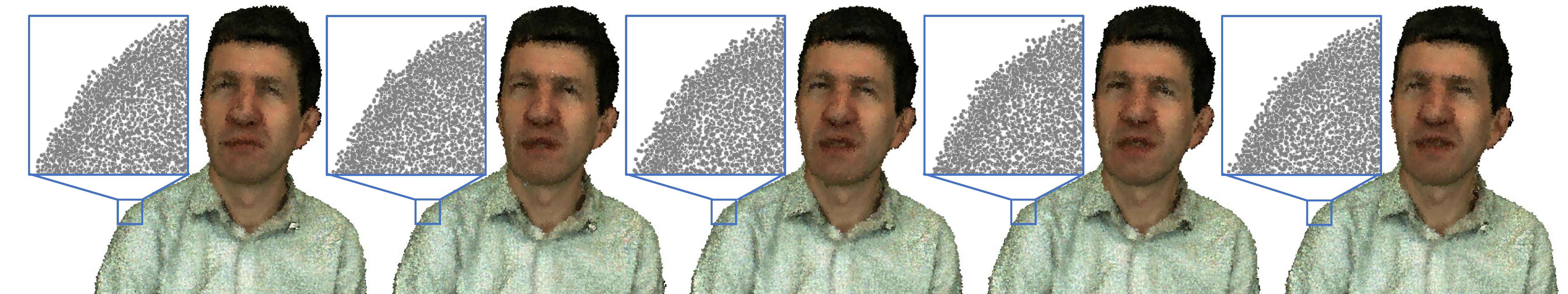}}
    \subfigure[LR]{\includegraphics[width=\textwidth]{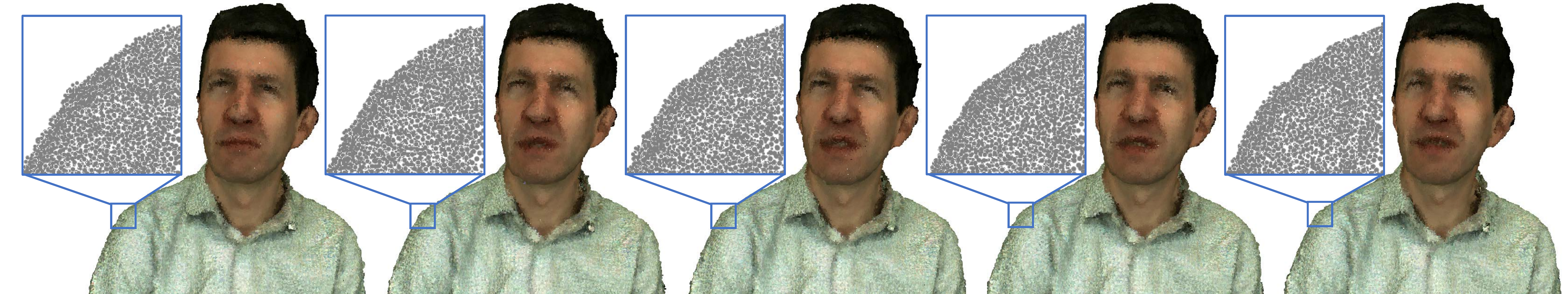}}
    \subfigure[MRPCA]{\includegraphics[width=\textwidth]{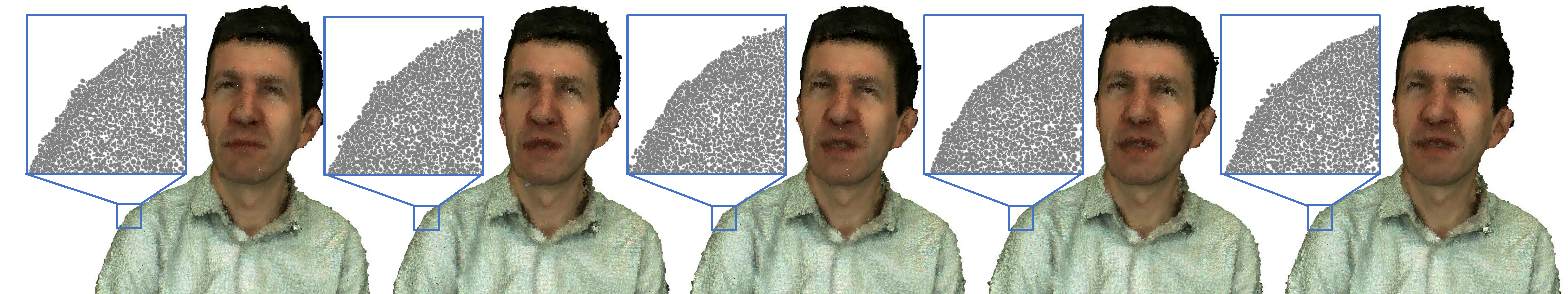}}
    \subfigure[Ours]{\includegraphics[width=\textwidth]{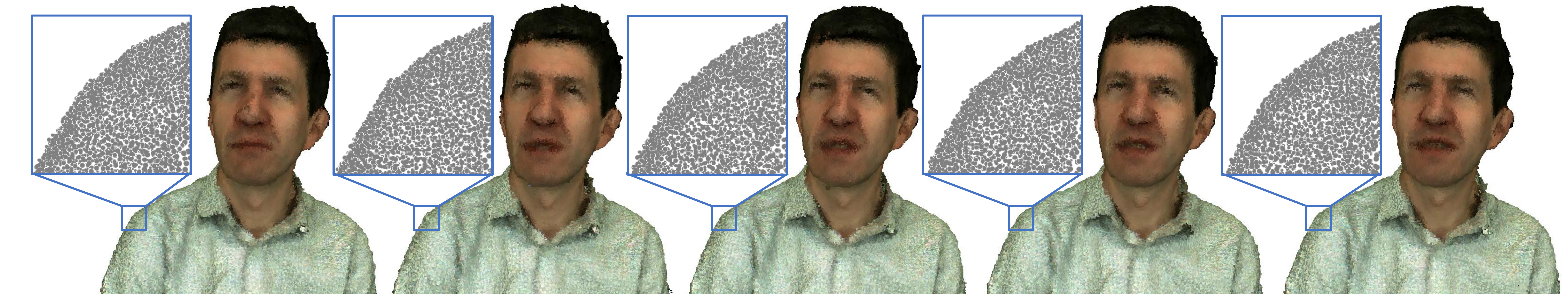}}
    \caption{Comparison results with Gaussian noise $\sigma=0.05$ for \textit{Andrew}: (a) The ground truth; (b) The noisy point cloud; (c) The denoised result by LR; (d) The denoised result by MRPCA; (e) The denoised result by our algorithm. Colors are not shown in the magnified regions for clear demonstration of geometry denoising.}
    \label{fig:andrew}
\end{figure*}

\begin{figure*}[htbp]
    \centering
    \subfigure[Ground-truth]{\includegraphics[width=\textwidth]{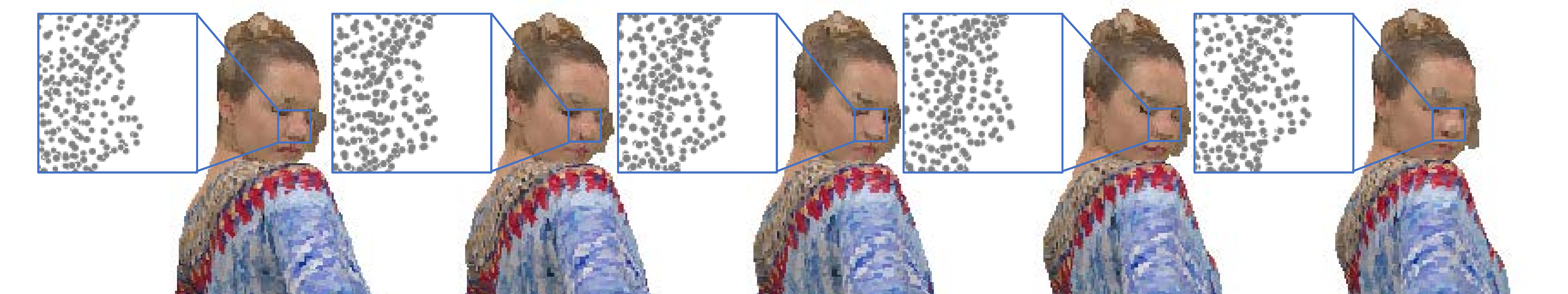}}
    \subfigure[Noisy]{\includegraphics[width=\textwidth]{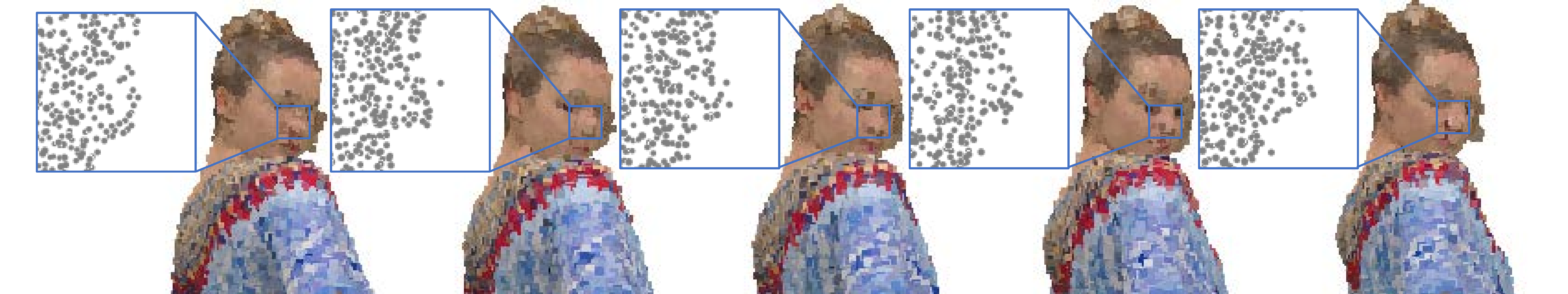}}
    \subfigure[LR]{\includegraphics[width=\textwidth]{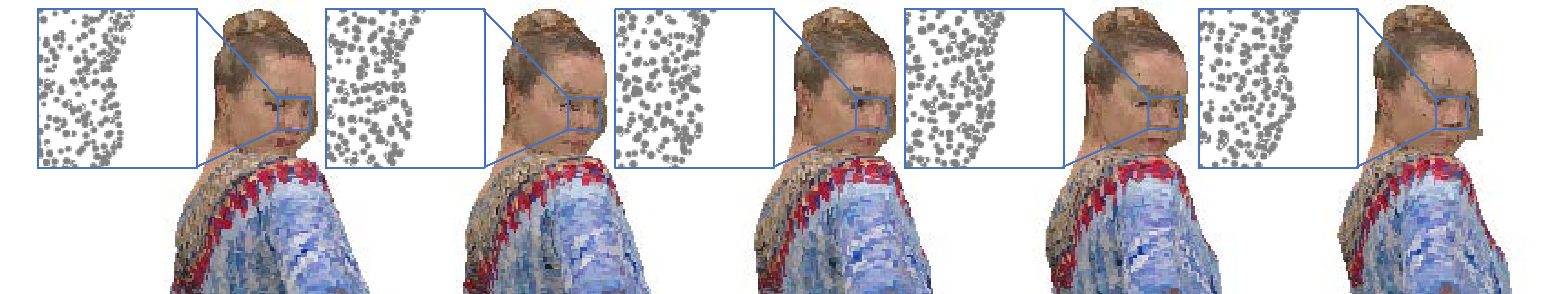}}
    \subfigure[MRPCA]{\includegraphics[width=\textwidth]{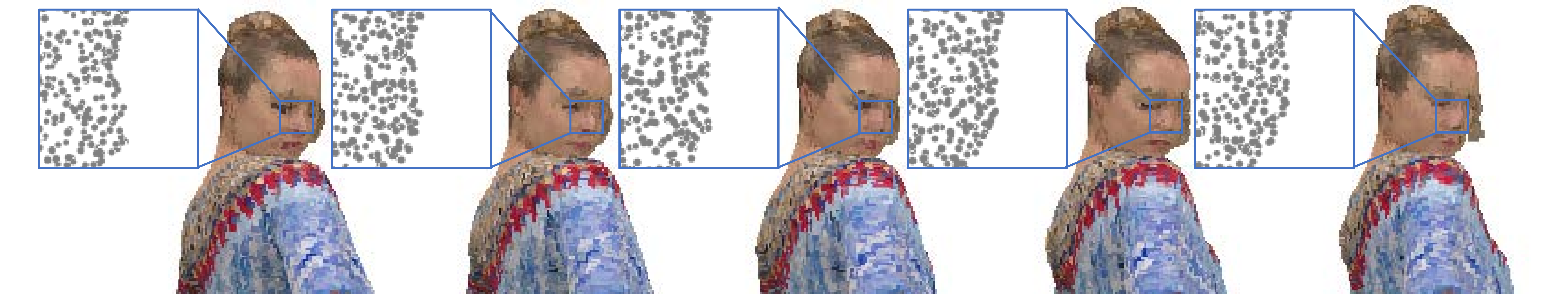}}
    \subfigure[Ours]{\includegraphics[width=\textwidth]{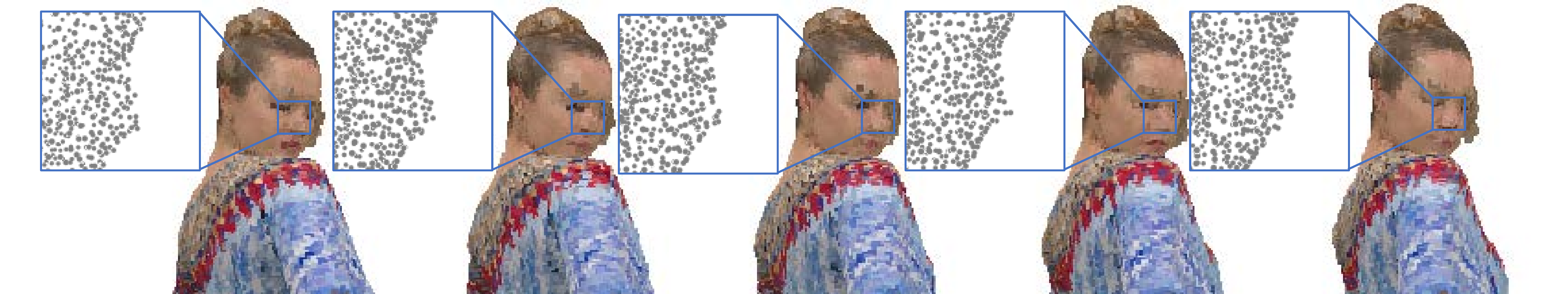}}
    \caption{Comparison results with Gaussian noise $\sigma=0.05$ for \textit{Longdress}: (a) The ground truth; (b) The noisy point cloud; (c) The denoised result by LR; (d) The denoised result by MRPCA; (e) The denoised result by our algorithm. Colors are not shown in the magnified regions for clear demonstration of geometry denoising.}
    \label{fig:long}
\end{figure*}

As illustrated in Fig.~\ref{fig:andrew} and Fig.~\ref{fig:long}, the proposed method also improves the visual results significantly, especially in local details and temporal consistency. 
In order to demonstrate the temporal consistency, we choose $5$ consecutive frames that exhibit apparent movement in $\textit{Andrew}$ from the MSR benchmark and $\textit{Longdress}$ from the MPEG benchmark, both under the noise variance $\sigma=0.05$. 
We show the visual comparison with MRPCA and LR because they are the nearest two competitors to our method in objective performance, as presented in Fig.~\ref{fig:compare_chart}. 

We see that our results preserve the local structure and keep the temporal consistency better. 
For instance, in the \textit{Andrew} model, while the results of LR and MRPCA still exhibit noisy contours even with outliers along the shoulder, our results are much smoother and cleaner. 
Also, note that, the ground truth of \textit{Andrew} is a little bit noisy due to the inherent limitation of the acquisition equipment---RGBD cameras, while our method is able to attenuate the sensor noise well.    
In \textit{Longdress}, while the nose of the model is over-smoothing and even distorted in shape in the results of LR and MRPCA, our method preserves the structure of the nose much better. 
Further, our results are much more consistent in the temporal domain, which validates the effectiveness of the spatio-temporal graph learning. 





\vspace{-0.05in}
\section{Conclusion}
\label{sec:conclude}
\vspace{-0.05in}
We propose 3D dynamic point cloud denoising based on spatio-temporal graph learning, exploiting both the intra-frame self-similarity and inter-frame consistency. 
Assuming the availability of a relevant feature vector per node on a point cloud sequence, we pose spatio-temporal graph learning as the problem of Mahalanobis distance metric learning, where we minimize the Graph Laplacian Regularizer and decompose the symmetric and positive-definite metric matrix for ease of optimization via proximal gradient. 
Then we formulate dynamic point cloud denoising as the joint optimization of the desired point cloud and underlying spatio-temporal graph, which is regularized by both intra-frame smoothness among searched similar patches and inter-frame consistency between corresponding patches.
Experimental results show that our method significantly outperforms independent denoising of each frame from state-of-the-art static point cloud denoising approaches. 

\vspace{-0.05in}

\end{document}